\documentclass{article}

\usepackage{microtype}
\usepackage{graphicx}
\usepackage{booktabs} 
\usepackage{graphicx}
\usepackage{subcaption}

\usepackage{hyperref}

\usepackage[accepted]{icml2023}

\usepackage{amsmath}
\usepackage{amssymb}
\usepackage{mathtools}
\usepackage{amsthm}
\usepackage{soul}
\usepackage{multirow}

\usepackage[capitalize,noabbrev]{cleveref}

\theoremstyle{plain}

\theoremstyle{definition}

\theoremstyle{remark}

\usepackage[textsize=tiny]{todonotes}
\usepackage{dsfont}

\DeclareMathOperator*{\argmin}{arg\,min}
\newcommand{\ind}[1]{\mathds{1}\left[ #1 \right]}

\icmltitlerunning{Text-To-Concept (and Back) via Cross-Model Alignment}

\begin{document}

\twocolumn[
\icmltitle{Text-To-Concept (and Back) via Cross-Model Alignment}



\icmlsetsymbol{equal}{*}

\begin{icmlauthorlist}
\icmlauthor{Mazda Moayeri}{equal,yyy}
\icmlauthor{Keivan Rezaei}{equal,yyy}
\icmlauthor{Maziar Sanjabi}{comp}
\icmlauthor{Soheil Feizi}{yyy}
\end{icmlauthorlist}

\icmlaffiliation{yyy}{Department of Computer Science, University of University of Maryland}
\icmlaffiliation{comp}{Meta AI}

\icmlcorrespondingauthor{Mazda Moayeri}{mmoayeri@umd.edu}
\icmlcorrespondingauthor{Keivan Rezaei}{krezaei@umd.edu}

\icmlkeywords{Machine Learning, ICML}

\vskip 0.3in
]



\printAffiliationsAndNotice{\icmlEqualContribution} 

\begin{abstract}
We observe that the mapping between an image's representation in one model to its representation in another can be learned surprisingly well with just a linear layer, even across diverse models. Building on this observation, we propose {\it text-to-concept}, where features from a fixed pretrained model are aligned linearly to the CLIP space, so that text embeddings from CLIP's text encoder become directly comparable to the aligned features. With text-to-concept, we convert fixed off-the-shelf vision encoders to surprisingly strong zero-shot classifiers for free, with accuracy at times even surpassing that of CLIP, despite being much smaller models and trained on a small fraction of the data compared to CLIP. 
We show other immediate use-cases of text-to-concept, like building concept bottleneck models with no concept supervision, diagnosing distribution shifts in terms of human concepts, and retrieving images satisfying a set of text-based constraints. Lastly, we demonstrate the feasibility of {\it concept-to-text}, where vectors in a model's feature space are decoded by first aligning to the CLIP before being fed to a GPT-based generative model. Our work suggests existing deep models, with presumably diverse architectures and training, represent input samples relatively similarly, and a two-way communication across model representation spaces and to humans (through language) is viable. 


\end{abstract}


\begin{figure}[htb]
\centering
    \includegraphics[width=\linewidth]{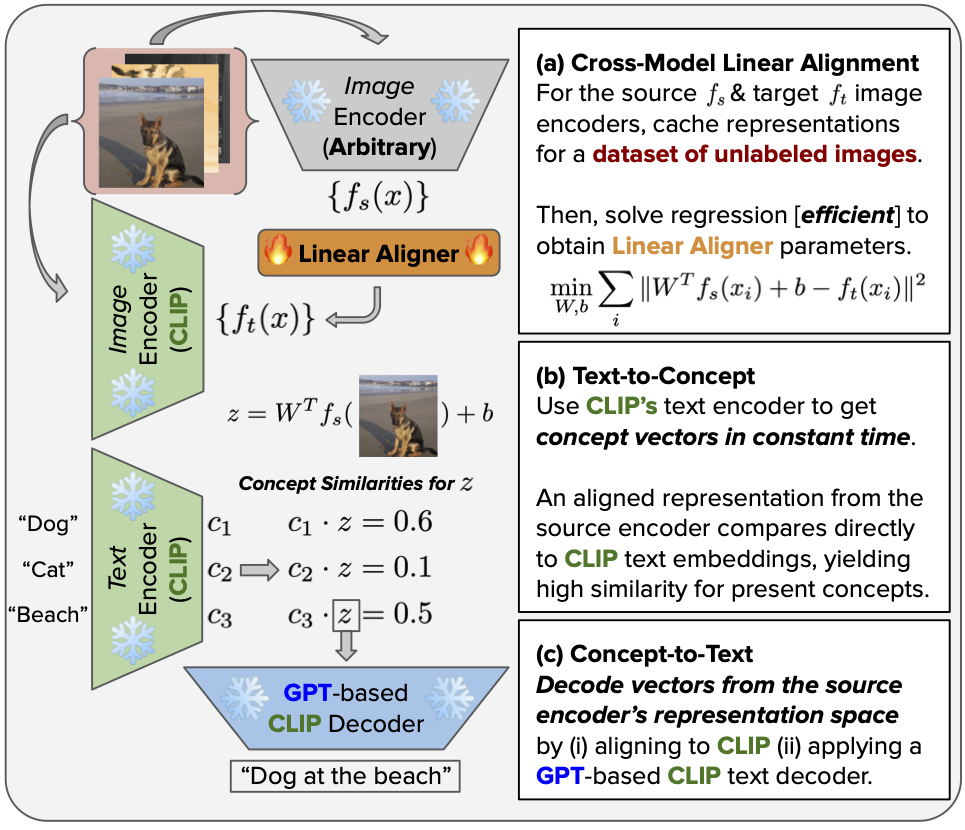}
    \caption{
        Overview. After \emph{aligning} the representation space of a given image encoder to a CLIP image encoder, we can compare aligned representations of images to concept vectors obtained \emph{directly from text} (typically, an example set of data is required to obtain each concept vector; our method is example-free, i.e. O(1) w.r.t. data collection).  Using a GPT-based CLIP decoder, we can map arbitrary vectors in representation space to text. Our method yields efficient \textbf{interpretability}: we only train \emph{one linear layer}.
    }
    \label{fig:intro-text-to-concept}
\end{figure}

\section{Introduction}
\begin{figure*}
    \centering
    \subfloat{\includegraphics[width=\textwidth]{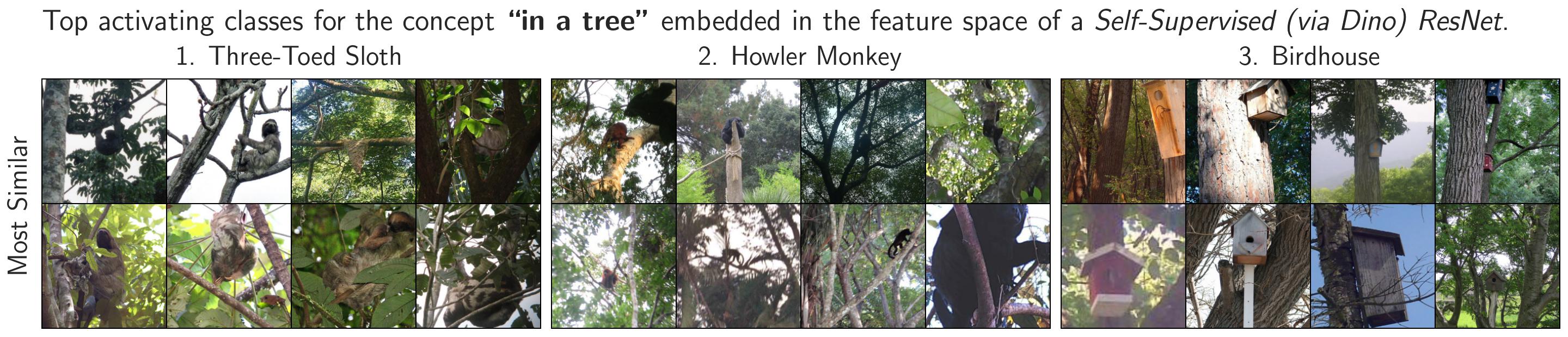}} \\
    \vspace{-0.5cm}
    \subfloat{\includegraphics[width=\textwidth]{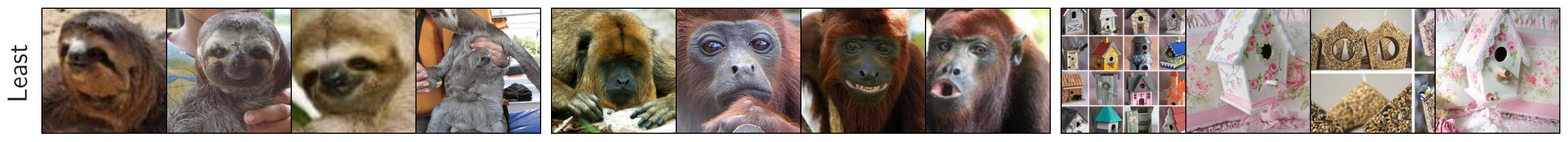}}\vspace{-0.1cm}
    \caption{Qualitative validation of text-to-concept. ImageNet classes are sorted by the average cosine similarity of the CLIP  embedding for ``in a tree'' to the \emph{linearly aligned} Dino ResNet representations of images within each class. Highest ranked classes indeed often appear in a tree, as is evident by the most similar instances. The least similar instances appropriately do not contain the concept.}
    \label{fig:eg_in_a_tree}
\end{figure*}

The representation spaces of deep vision models are undoubtedly rich in semantic structure. However, these deep feature spaces are notoriously challenging for humans to interpret, mainly because it is hard for us to digest thousands of numbers at once. Unlike deep models, which encode concepts as vectors in high (e.g. $d=2048$) dimensional spaces, humans have developed language to describe the world around us concisely. In this work, we propose a method to map text to concept vectors that can be compared directly to image representations obtained from off-the-shelf vision encoders trained with {\it no text supervision}.

Our method works by {\it aligning} the representation space of a given vision model to the representation space of a CLIP \cite{Radford2021LearningTV} model. By design, the CLIP representation space is shared across jointly trained vision and text encoders. Thus, CLIP models already have text-to-concept built in, via the text encoder. To extend this capability to off-the-shelf models, we propose to learn a mapping between representation spaces. Specifically, we optimize a function to predict the representation of an image for a target model (i.e. CLIP) from the same image's representation for a source model (i.e. off-the-shelf vision model). We can then map the representations of the off-the-shelf model to CLIP space, where the aligned features would reside in the same space as the concept vector for the desired text. 

The mapping function, however, may significantly change the semantics of its input. To prevent this, we restrict the hypothesis space of our mappings to be {\it affine transformations}. Despite their simplicity, we find that linear layers are surprisingly effective at performing feature space alignment, even between models with diverse architectures and training procedures. This observation suggests that despite drastically different approaches to training, diverse models seem to learn to store information in similar ways. Most notably, we can align model representations to CLIP, thus extending CLIP's text-to-concept abilities to existing models. 

\looseness -1
Figure \ref{fig:eg_in_a_tree} visually validates our approach: after encoding the concept ``in a tree'' in CLIP space and computing similarity with aligned representations from a self-supervised ResNet, the classes with the highest average similarity are reasonable, and images within them with the highest similarity prominently display the concept, while the least similar class instances do not. Stronger validation of our approach is found in performing {\bf zero-shot classification using off-the-shelf encoders via text-to-concept.} Models achieve impressive zero-shot accuracy on many tasks, often being competitive with a CLIP model that is larger, trained on many more samples with richer supervision, and most notably, directly optimized to align with the text encoder we use in text-to-concept. Surprisingly, zero-shot accuracy of off-the-shelf models surpasses the CLIP in a few cases, particularly for color recognition. While greatly expanding the use cases for existing models, these zero-shot abilities also support the belief that deep models learn many more abstract notions than what they are explicitly trained to know. Text-to-concept allows for uncovering and better utilizing the rich semantics hidden in existing models' representation spaces.

\looseness -1
In addition to zero-shot learning for free, text-to-concept has several immediate interpretability applications, such as converting vision encoders to {\it Concept Bottleneck Models} (CBMs) \cite{cbm} with {\it no concept supervision required}. CBMs decompose inference into a concept prediction step followed by class prediction using a white box model (i.e. linear head) on concept predictions, so that the contribution of each concept to the final logit can be precisely computed. Typically, CBMs require concept supervision in addition to class labels, but with text-to-concept, we can replace concept predictions with concept similarities, obtained by comparing aligned representations to the vector obtained for the desired notion. Then, CBM training reduces to simply training a linear layer to predict class labels from pre-computed similarities of aligned image representations to a set of concept vectors. We illustrate this application on RIVAL10 data \cite{rival10}, which has attribute labels, though we only use these labels to verify our zero-shot concept prediction approach. Indeed, we obtain an AUROC of $0.8$ for RIVAL10 attribute prediction in the zero-shot manner described, leading to a highly accurate ($93.8\%$) resultant CBM with desired interpretability benefits (see Figure~\ref{fig:cbn_eg_frog}).  

\looseness -1
Next, we show text-to-concept can demystify large datasets, as the distribution of similarities between a bank of text-to-concept vectors and aligned representations of the data essentially summarizes what concepts are present, explaining the data distribution in human terms. This can be applied to diagnosing distribution shifts, as we can inspect the shift w.r.t. to human-understandable concept similarities. For example, when comparing ImageNet to ObjectNet \cite{objectnet}, we can show that the distribution of similarities for the ``indoors'' concept shifts dramatically, capturing the essence of why ObjectNet poses a challenge: images in ObjectNet were taken in people's homes. Another way text-to-concept aids in engaging with large datasets is via concept-based image retrieval. Using \emph{concept logic}, we query the image representations for a given model that satisfy a set of concept similarity thresholds, allowing for greater human control of the importance of each concept in the search, yielding reasonable results in finding specific images in a large corpus. 

Lastly, we close the human-machine communication loop by introducing \emph{concept-to-text} to directly decode vectors in a model's representation space. Our implementation aligns the model's space to CLIP, and then leverages an existing CLIP space decoder \cite{zerocap} that uses a CLIP embedding to guide the output of GPT-2. The existing decoder was intended for image captioning, though we demonstrate that its abilities extend to general vectors (i.e. not obtained from a single image) from non-CLIP models after alignment. Specifically, we decode the vectors in the classification head of three ImageNet trained models. We then use a human study to verify that the decoded captions describe the class associated with each vector, finding that our simple method works in over $92\%$ of cases.

Our methods extend the capabilities of multi-modal models like CLIP to other models that are trained on much smaller uni-modal datasets and with weaker supervisions. This can be useful when a model more accurate or smaller than CLIP for a specific domain is desired, or when training CLIP-like models is infeasible, due to the large corpus of image-caption pairs needed. Moreover, since our approach can be applied to interpret any model's representation space, while only requiring the training of a linear layer, its potential impact is very high, as text-to-concept is easy to plug-in and has a breadth of applications. The implications of our work are also startling: first, the success of linear representation space alignment indicates that diverse models ultimately represents inputs relatively similarly. The emergent zero-shot abilities of existing models suggests an under-utilization of models we already have. Finally, the synergy we display between CLIP, GPT, and existing models, coupled with the ability to communicate across these models and back and forth with humans, makes the prospect of diverse models collaborating with minimal tuning very promising.

\section{Review of Literature}
Our alignment of model representation spaces is related to stitching, first introduced by \citet{lenc2015understanding}, who train linear layers to merge top and bottom chunks of different models, resulting in ``franken-CNNs''. Stitching was revisited by \citet{bansal2021revisiting}, who aimed to showcase how it can be used as a tool to quantify the quality of representations towards learning how to obtain better representations, and \citet{csiszarik2021similarity}, who consider different ways to train stitching layers. We note these works typically stitch together models of the same architecture, where as we consider a much more diverse set of models. Also, those works focus on comparing representations to one another, while we aim to relate representations to human notions.

Namely, we seek to obtain concept vectors within the representation space of off-the-shelf models from text. Also known as concept activation vectors (CAVs), \citet{kim2018interpretability} popularized the study of directions corresponding to human concepts in deep feature spaces, as well as the sensitivity of model outputs to changes along these directions, so to interpret deep networks. One limitation is the necessity of example sets of data to define CAVs. More recent efforts automatically discover CAVs \cite{ACE, craft, Zhang2020InvertibleCE}, though annotating the discovered concepts with language is not straightforward, which motivates our concept-to-text method. 

The rise of joint vision-language models like CLIP \cite{Radford2021LearningTV} make it possible to interpret vision space with text, as well as perform zero-shot classification. Follow up works leverage CLIP to annotate neural nodes \cite{oikarinen2022clip} or to distill failure modes \cite{jain2022distilling} of non-CLIP models. However, they engage probe datasets or exemplars to communicate with CLIP space, while our method directly aligns representation spaces. Recently,  \citet{Moschella2022RelativeRE} devise a zero-shot method for communication across representation spaces based on relative positions to anchor points, which supports our claim that representation spaces are sufficiently equivalent to where the extension of CLIP's inherent text-to-concept ability to off-the-shelf models via linear alignment is viable.

\newcommand{\Dtrain}{D_{\text{train}}}
\newcommand{\xtrain}{x^{\text{train}}}
\newcommand{\ytrain}{y^{\text{train}}}

\newcommand{\Dtest}{D_{\text{test}}}
\newcommand{\xtest}{x^{\text{test}}}
\newcommand{\ytest}{y^{\text{test}}}

\newcommand{\ftrans}{f_{s \rightarrow t}}
\newcommand{\htrans}{h_{s \rightarrow t}}
\newcommand{\Rsquare}{R^2}
\newcommand{\pc}{\text{pc}}
\newcommand{\diag}{\text{diag}}

\section{Model Alignment}
\label{sec:alignment}

We use $\mathcal{X}$ to denote the set of all possible input images.
Let $\Dtrain, \Dtest \subset \mathcal{X}$ denote the training and test datasets. We define a \emph{vision encoder} as a model $f$ that maps images $x\in \mathcal{X}$ to vectors $f(x)\in \mathbb{R}^d$.
Given two vision encoders $f_s, f_t$,
\emph{representation space alignment} of model $f_s$ to model $f_t$
is the task of learning a mapping $h:f_s(\mathcal{X})\rightarrow f_t(\mathcal{X})$.
We restrict $h$ to the class of affine transformations, i.e., $h_{W, b}(z) := W^T z + b$.

To maximally retain the original semantics of representation spaces,
we design the following optimization problem
\begin{align*}
    \label{eq:gen-alignment}
    W, b = \argmin_{W, b} \frac{1}{|\Dtrain|} \sum_{x \in \Dtrain} \|W^T f_{s}(x) + b - f_{t}(x)\|_2^2
    .
\end{align*}

The above optimization can be viewed as multiple linear regression problems;
thus we evaluate the linear alignment on $\Dtest$ by considering the quality of the solution on those linear regression problems.
We use \emph{Coefficient of Determination}, i.e., $\Rsquare$ which is
the proportion of the variation in the dependent variables that is predictable from the independent variables.
Furthermore, we note that for the vision encoder $f_s$,
there usually exists a \emph{classification head} $g_s: \mathbb{R}^{d} \rightarrow \mathcal{C}$
that classifies a representation in the space of model $f_s$.
Indeed, the predicted label for input $x$ is $g_s\left(f_s(x)\right)$.
Note that $\mathcal{C}$ denotes the set of labels, e.g., ImageNet classes.
We define \emph{aligned accuracy} as the accuracy of classification on $\Dtest$
when we use $f_s$ as the vision encoder, then do the linear transformation to obtain the corresponding representation in space of $f_t$, and finally, use $g_t$ for classification.
If alignment works well, aligned accuracy should be admissible and comparable to the accuracy of model $f_t$ when no alignment is used. Then, we define \emph{retained accuracy} as the ratio of aligned accuracy to the accuracy of model $f_t$ without any alignment.
Note that we use ImageNet-1K train and test datasets as $\Dtrain$ and $\Dtest$ in linear alignment.

\begin{figure}[tbp]
\centering
    \includegraphics[width=\linewidth]{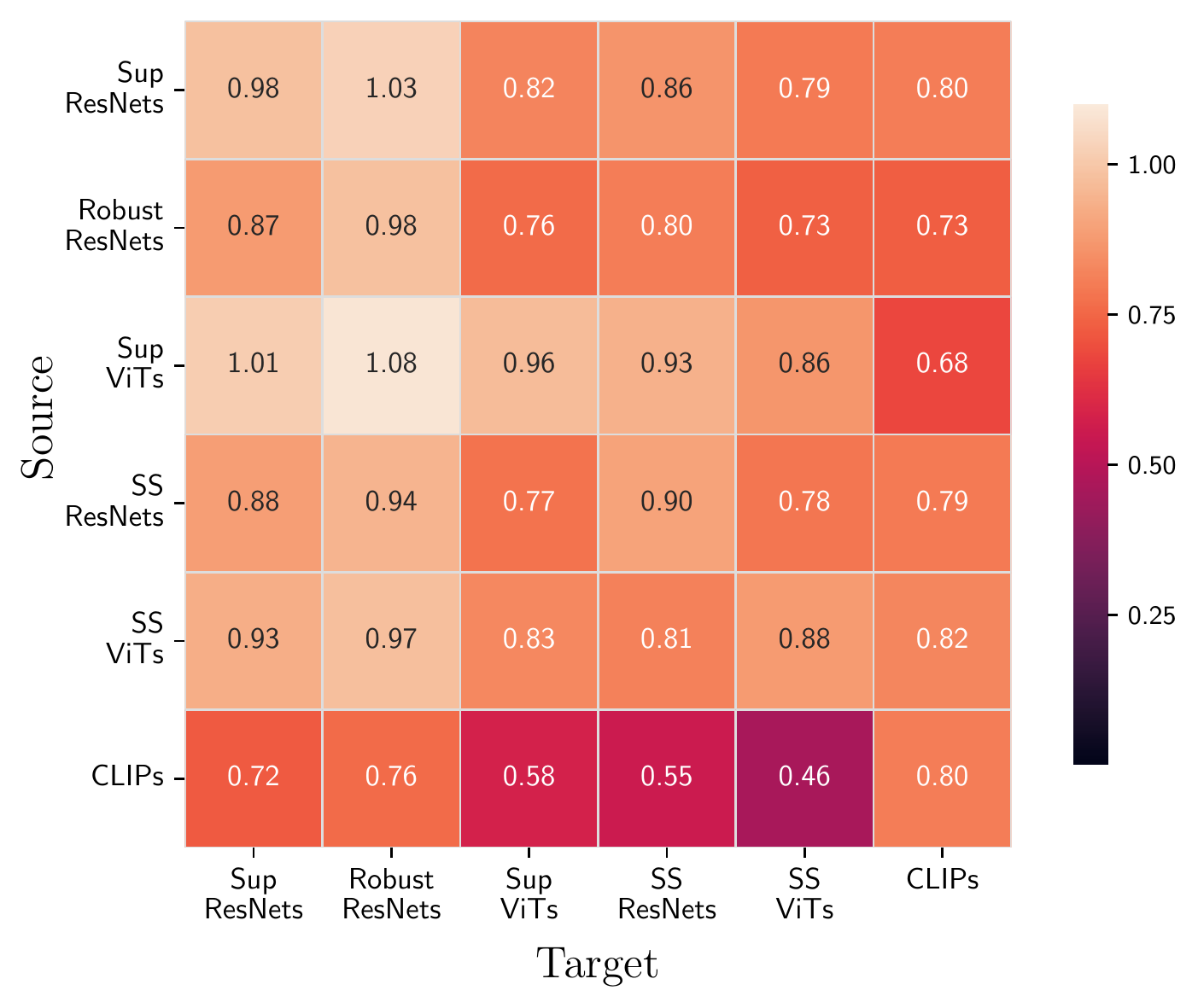}
    \caption{
    Heatmap of retained accuracy. 
    the value in row $r$ and column $c$ is the average of retained accuracy when doing alignment from all models in group $r$ to all models in group $c$.
    Note that ``Sup" refers to supervised while ``SS" refers to self-supervised models.
    All models except CLIPs are trained on ImageNet.
    }
    \label{fig:gen-alignment-main}
\end{figure}

Interestingly, we observe that simple linear alignment works well in terms of both $\Rsquare$, and aligned accuracy across various models. Figure~\ref{fig:gen-alignment-main}
shows the aligned accuracy between diverse pairs of models. 
In the scope of linear alignment, we further consider the sample efficiency of optimizing linear alignment as well as investigating linear alignment in space of top principal components of representation spaces, where we roughly see strong, near identity correspondence between the top principal components of different models. We refer to Appendix~\ref{app-sec:alignment} and \ref{app-sec:pca} for more details.




According to Figure~\ref{fig:gen-alignment-main}, various models are highly alignable to CLIP models. This is surprising as CLIP models are trained on other datasets than ImageNet and their training procedure involves vision/text supervision which is drastically different from other models. High-quality alignment to CLIP representation space enables models to adopt a wide variety of CLIP models capabilities, which we analyze in this work.
On the other hand, we observe that retained accuracy when aligning CLIP models to other models is not high. This is mainly due to the fact that CLIP models encode images and texts in relatively low-dimensional spaces and in linear regression, approximating dependent variables becomes harder as the number of independent variables decreases. Indeed, linear alignment works worse when we align from a representation space with lower dimensionality to a representation space with higher dimensionality.

\section{Text to Concept}
\begin{figure*}[htb]
    \centering
    \subfloat{\includegraphics[width=\textwidth]{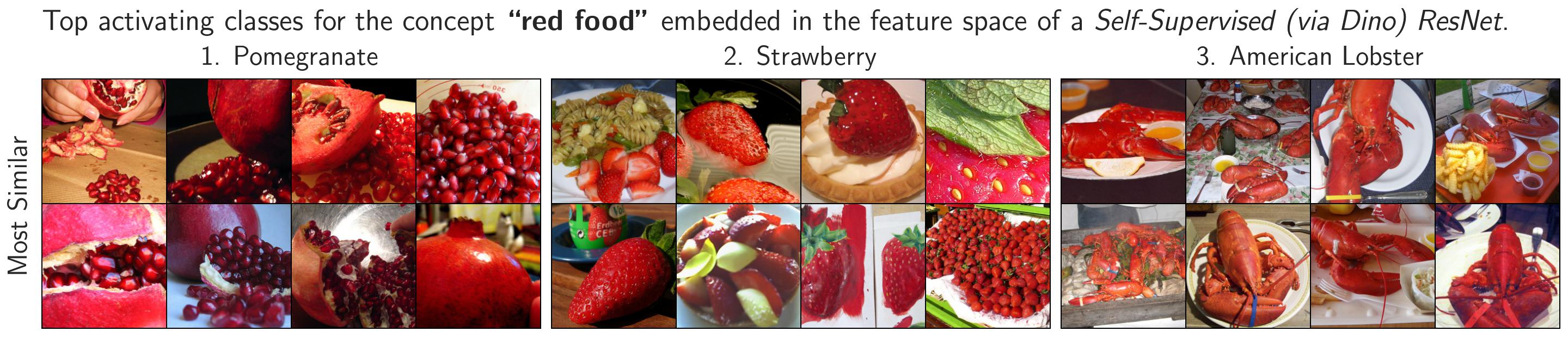}} \\ 
    \vspace{-0.5cm}
    \subfloat{\includegraphics[width=\textwidth]{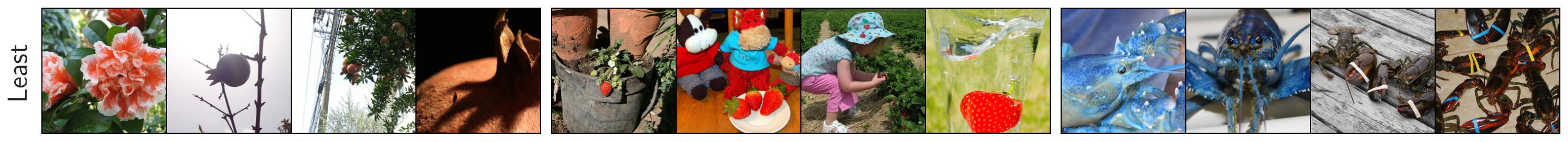}} \\
    \vspace{-0.1cm}
    \centering
    \subfloat{\includegraphics[width=\textwidth]{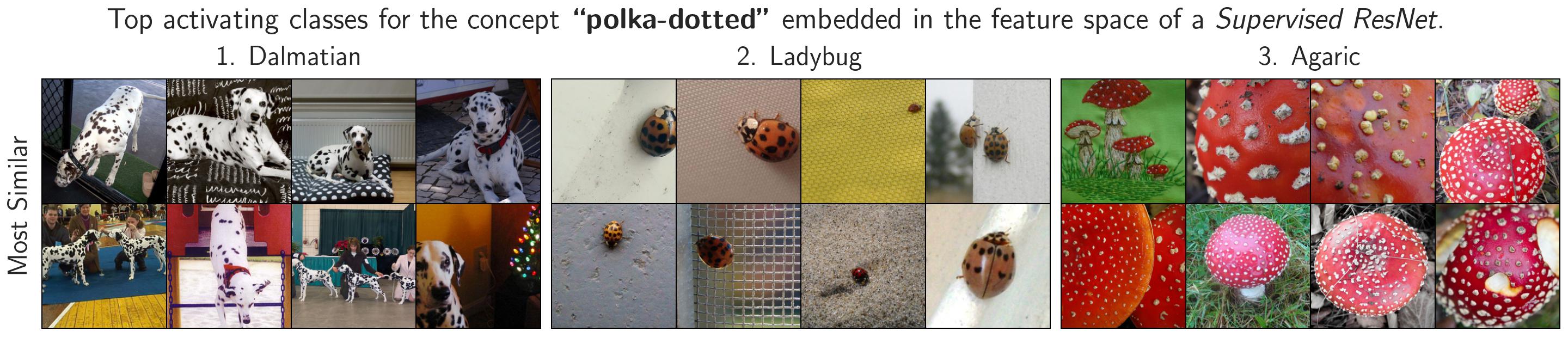}} \\ 
    \vspace{-0.5cm}
    \subfloat{\includegraphics[width=\textwidth]{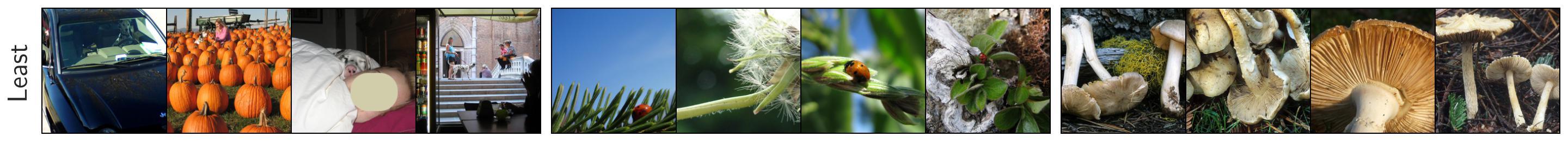}}\vspace{-0.1cm}
    \caption{Text-to-concept can encode finer-grain concepts, like combinations of concepts (\emph{``red food''}) or textures (\emph{``polka-dotted''}).} 
    
    \label{fig:qual_val}
\end{figure*}


Leveraging representation space alignment, specifically to CLIP, we perform text-to-concept, where text descriptions of semantic concepts are encoded as vectors that can be directly compared (i.e. via cosine similarity) with the aligned features of images obtained from an off-the-shelf vision encoder (see Figure \ref{fig:intro-text-to-concept}). 
Despite its simplicity, alignment-based text-to-concept is surprisingly effective, which, after further detailing our method, we demonstrate qualitatively and quantitatively in this section. Notably, we show that the similarities of aligned image representations to class vectors obtained via text-to-concept enables \emph{zero-shot classification for non-CLIP models off-the-shelf}, with zero-shot accuracy of much simpler models at times exceeding that of CLIP.

\subsection{Method Details}

We define text-to-concept as a procedure for obtaining vectors corresponding to concepts described as text that can be directly compared (i.e. via cosine similarity) to image representations from a fixed vision encoder. Our method begins with a string describing some concept, like ``red food''. We then prepend this string with a number of template prompts (e.g. ``a photo of \{\}''); we use the same template prompts as in CLIP's original paper for ImageNet zero-shot classification. Then, we embed the templated text to CLIP space using CLIP's text encoder, and average the resultant vectors over all templates to obtain a single concept vector (as is standard). For some object agnostic concepts, such as contexts like ``in a tree'', we can encode a general prompt like ``a photo of an object in a tree'', or we can obtain a more refined vector by encoding ``\{\emph{prompt}\} \{\emph{class name}\} in a tree'', averaging over all choices for \emph{class name} and \emph{prompt}. There are countless ways to prompt engineer; we elect to use general prompts in most cases, as prompt engineering is not the focus of our work.\footnote{See Appendix \ref{app-sec:ttc} for complete details on all prompts used.} 

\begin{figure*}[htb]
    \centering
    \subfloat{\includegraphics[width=\linewidth]{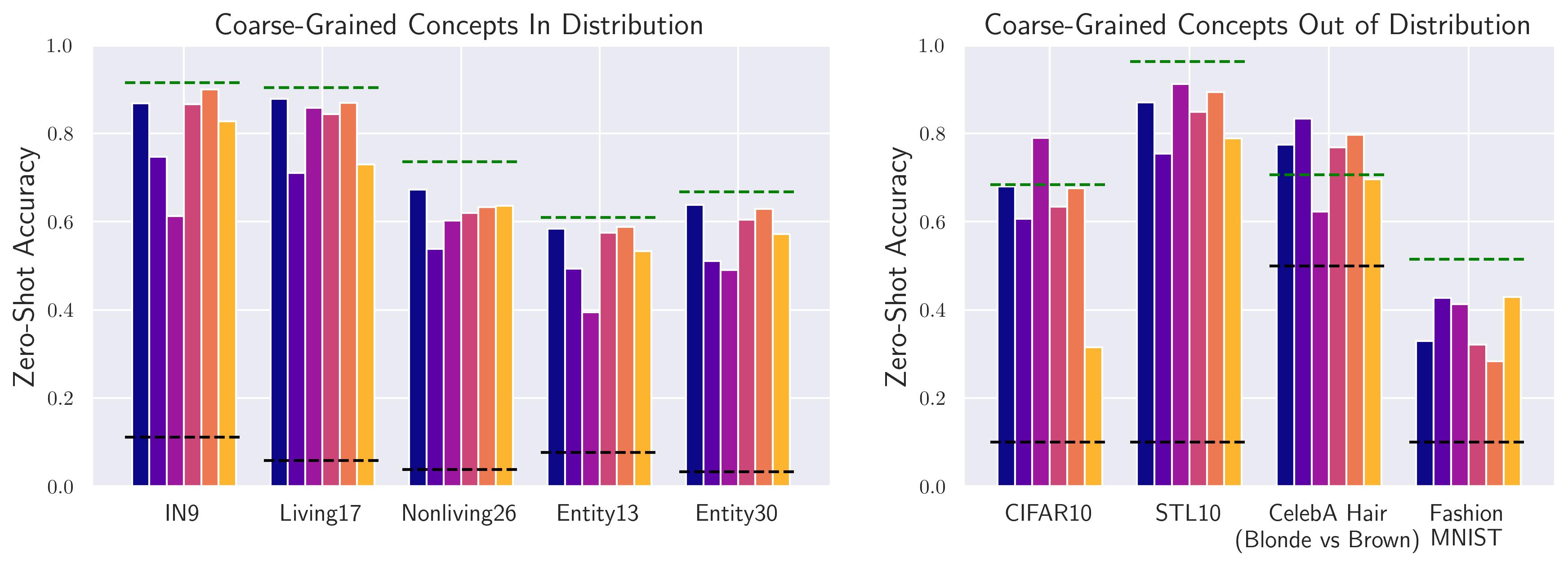}} \\
    \vspace{-0.1cm}
    \centering
    \subfloat{\includegraphics[width=0.7\textwidth]{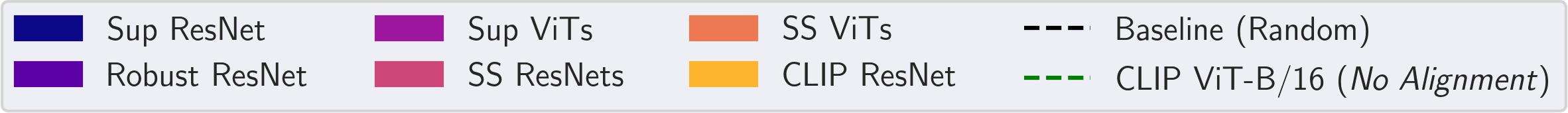}}
    \caption{The zero-shot capabilities of CLIP can extend to off-the-shelf vision encoders via alignment based text-to-concept. ({\bf Left}) Models trained on ImageNet can recognize coarse categorizations of ImageNet classes, despite never explicitly being taught them. ({\bf Right}) Off-the-shelf models remain strong zero-shot classifiers even when images are out of distribution. In some cases, they surprisingly surpass the accuracy of the CLIP vision encoder whose jointly-trained text encoder was used to embed each class vector.}
    \label{fig:zero_shot}
\end{figure*}

Then, for a given model, we train a linear layer to align its representation space to CLIP; specifically, we use CLIP ViT-B/16. 
We pass ImageNet training images to the given model's feature encoder and CLIP's vision encoder, resulting in a dataset of paired representations with which we train our aligner (Section \ref{sec:alignment}). 
Now, we have two functions that map to CLIP's vision space: the CLIP text encoder (since the text and vision representation spaces are shared), and the composition of the given model's encoder with the linear aligner. Since the concept vector obtained via CLIP's text encoder and aligned representations from the given model are both mapped to the same space, we can compare them directly, thus satisfying our definition of text-to-concept. Alternatively, we could train an aligner from CLIP to the given model's representation space, and align the text embedding instead of the features. We found this method to be less effective, possibly because the dimensionality of CLIP space is lower than most models we study. Since our aligner is a simple affine transformation, alignment minimally changes the content of the representation obtained from the off-the-shelf model. Also, note the efficiency of our approach: after training a linear layer once, we can encode any number of new concepts from text at no additional training cost. 


\looseness=-1
{\bf Qualitative Validation}: Figures \ref{fig:eg_in_a_tree} and \ref{fig:qual_val} show images selected based on the cosine similarity of their aligned representations (obtained using off-the-shelf encoders and trained linear aligners) to certain concept vectors. For each concept, we present the classes with the highest average similarity, as well as the most and least similar images within them. The retrieved classes are sensible for each concept (e.g. \emph{American Lobster} for ``red food''). Sorting images within each class separates examples where the concept is extremely prominent from those where the concept is absent (e.g. images of uncooked lobsters are least similar to the ``red food'' concept). Note that the models used to obtain the image representations differ in architecture, training objective and supervision from CLIP, and most notably, they have not been trained with any text/concept supervisions. Thus, it is surprising that we can easily connect these visual concept representations to the CLIP text embeddings.
Nonetheless, over a range of concept types (a context, a combination of color and a concept, and a texture), our visualizations qualitatively validate our proposed text-to-concept approach. We now turn to zero-shot classification for quantitative validation.

\subsection{Zero-Shot Classification}
\label{sec:zero_shot}

CLIP models perform zero-shot classification by comparing image representations to embeddings of text strings describing each class: the predicted class is the one whose text embedding is most similar to the test image's representation. This is referred to as zero-shot since no labeled instances from the candidate classes are used. Considering classes as concepts, we can then use text-to-concept to obtain vectors that are directly comparable to aligned representations from off-the-shelf vision encoders, thus extending CLIP's zero-shot capabilities. The accuracy of zero-shot classification serves as a quantitative measure of the quality of text-to-concept vectors. Indeed, when concept vectors align better with representations of samples in the class, zero-shot accuracy is higher. Thus, we explore zero-shot classification over many datasets to shed insight on when and how well text-to-concept works. We consider models over diverse architectures and training procedures, though all models are roughly equal in size ($\sim$25M parameters) and are only trained on ImageNet (except for CLIP). Also, the baseline CLIP model (ViT-B/16) whose text encoder is used to embed concepts is much larger in size ($\sim$80M parameters); this baseline is intended more so as an upper bound. 

First, we ask if models can recognize new categorizations of the data they were trained over. Namely, we consider coarse grained categorizations of ImageNet classes (e.g. distinguishing {\it insects} from {\it carnivores}, see \cite{in9, breed} and Appendix \ref{app-sec:zero_shot}). We also investigate if these coarse grained concepts can still be recognized as image data is taken out of distribution. Figure \ref{fig:zero_shot} displays the results. We observe impressive zero-shot performance in both cases. For example, on a 17-way classification problem, self-supervised ViTs achieve $85\%$ accuracy, despite never receiving supervision about these classes, or any classes at that. Shockingly, in a few cases, even the performance of the CLIP model whose text encoder (with which it was jointly trained) was used to obtain concept vectors is surpassed.

\begin{figure}[htb]
    \centering
    \includegraphics[width=\linewidth]{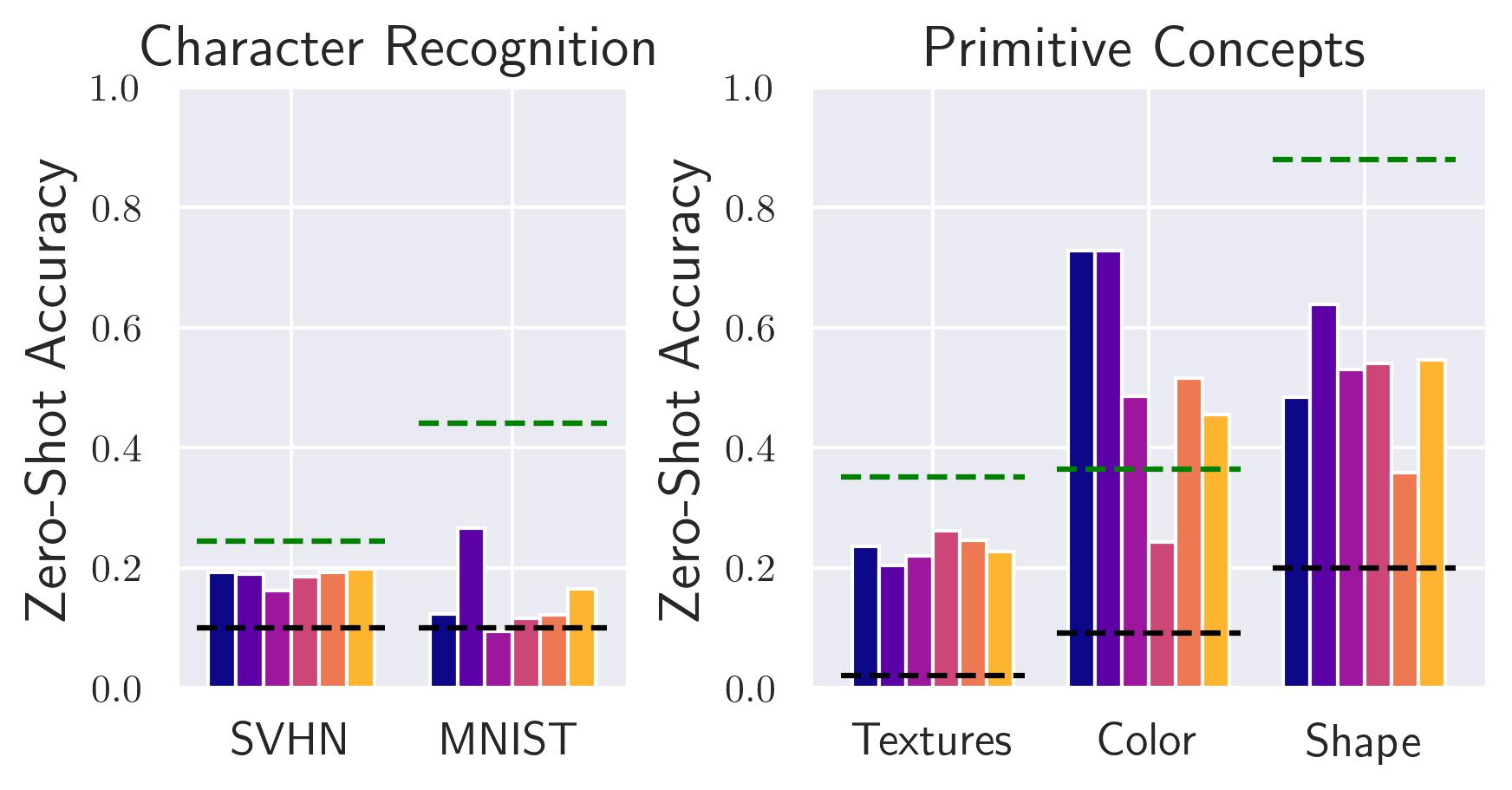}
    \caption{Edge cases for zero-shot classification. ({\bf Left}) Models struggle with OCR. ({\bf Right}) Models can recognize some primitive concepts by name. Same legend as figure \ref{fig:zero_shot}.}
    \label{fig:zero_shot_edge_cases}
\end{figure}

To stress test text-to-concept, we consider tasks that require models to recognize characters (specifically digits) or primitive concepts, like textures, colors, and shapes. We observe most models only marginally surpass random accuracy for character recognition tasks. Oddly, the adversarially trained ResNet is roughly twice as good as other models in zero-shot MNIST classification, though it still performs far worse than the baseline CLIP model, which also struggles. This suggests models simply may not have any notion as to what distinguishes digits from one another, which is not surprising given that it would not be very useful for understanding ImageNet images. On the other hand, models achieve far better than random performance in recognizing primitive concepts, which appear in ImageNet as low level features for more abstract notions. Interestingly, color recognition is a task where most models outperform the CLIP baseline, suggesting the CLIP ViT may have reduced color sensitivity relative to other primitive concepts. 

While these experiments validate our proposed text-to-concept method, it is also remarkable that these off-the-shelf models, who have much smaller training sets (roughly $0.3\%$ the size of CLIP's) and receive far less supervision, are comparable to CLIP in recognizing the unseen classes we consider. This suggests that models learn far more than what they are taught. In other words, models discover many semantic concepts and organize their representation spaces so that these concepts are roughly linearly separable, even when they are only explicitly directed to separate $1000$ classes or to simply draw representations of similar inputs close to another. Thus, even models trained with elementary techniques likely contain much richer representation spaces than their use case requires. The success of transfer learning supports this claim, as a small amount of labeled data is sufficient for a model to recognize `new' concepts, implying they had some notion of the concepts before. Text-to-concept can enable better understanding and utilization of these rich representation spaces, \emph{without requiring new labeled samples}. 

\section{Additional Applications of Text-to-Concept} 


\subsection{Concept-Bottleneck Networks for Free}

\begin{figure}[htb]
    \centering
    \includegraphics[width=0.8\linewidth]{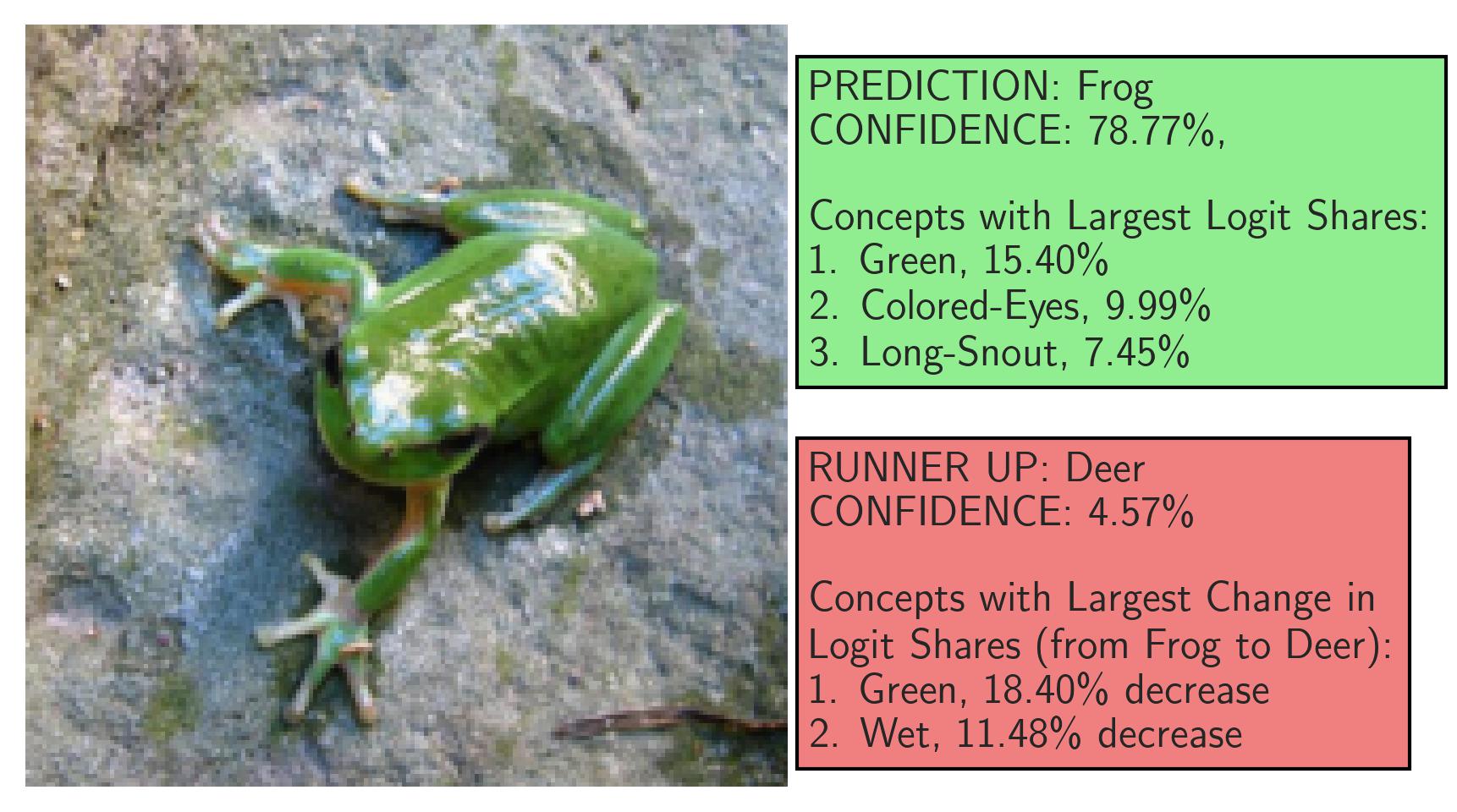}
    \caption{Example inference for a Concept Bottleneck Model (CBM) obtained via training a linear layer on zero-shot concepts. Since logits in the CBM are linear functions of concept scores, we can precisely quantify the contribution of concepts to each logit.}
    \label{fig:cbn_eg_frog}
\end{figure}

The zero-shot results suggest models are already aware of many concepts beyond those which they are directly trained to learn. One case where knowledge of concepts related to the classification task is salient is Concept Bottleneck Models (CBMs) \cite{cbm}. CBMs are intepretable by design, as they first predict the presence of concepts using a black box, and then obtain class logits with a white box (e.g. linear layer) atop concept predictions. Thus, the contribution of each concept to the predicted logit can be computed directly, allowing predictions to be faithfully explained with semantic reasons. A major barrier to using CBMs is that they require concept supervision, which can be prohibitively expensive. Text-to-concept, however, alleviates this constraint, thanks to zero-shot concept prediction. 

We use RIVAL10 classification \cite{rival10} as an example for how a CBM can be implemented with \emph{no concept supervision} using text-to-concept. RIVAL10 is an attributed dataset, though we do not use these labels during training. We use RIVAL10 because a linear classifier operating on ground truth attribute labels achieves $94.5\%$, indicating that a CBM could be effective. Further, the attribute labels allow for quantifying the quality of the zero-shot concept vectors we obtain. 

To implement the network, we use text-to-concept to encode the $28$ attributes annotated in RIVAL10 as vectors in CLIP space. We then compute the similarities between the attribute vectors and aligned (to CLIP) features from an ImageNet pretrained ResNet-50. Finally, we fit a linear layer atop image-attribute similarities (i.e. in representation space) to predict class labels. Note that the only training we conduct is that of the final classification head and of the aligner, both of which are linear layers, making them time and sample efficient to optimize. The resultant CBM achieves $93.8\%$ accuracy, and yields the desired interpretability advantages, as shown in figure \ref{fig:cbn_eg_frog}. Moreover, using image-attribute similarities (via text-to-concept) as a score for predicting attributes achieves an AUROC of $0.8$, with $72\%$ of attributes achieving at least an AUROC of $0.75$. Thus, zero-shots concepts are relatively accurate in predicting RIVAL10 attributes. See appendix \ref{app-sec:cbn} for details.


\subsection{Concept-Based Dataset Summarization and Distribution Shift Diagnosis}

\begin{figure}
    \centering
    \includegraphics[width=0.8\linewidth]{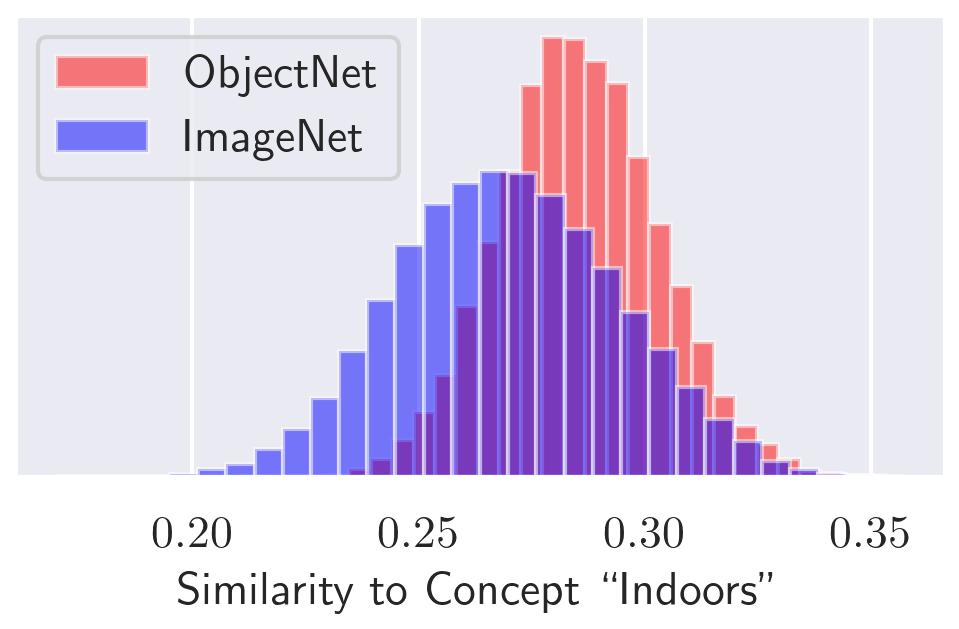}
    \caption{Concept similarities can reveal distribution shifts, like in ObjectNet, where photos are taken within people's homes.}
    \label{fig:objnet_vs_imgnet}
\end{figure}

The interpretability benefits of text-to-concept also apply to demystifying large datasets. Specifically, one can discern the presence of a concept in their data by using text-to-concept to obtain a corresponding vector, and computing the similarity of this vector to all aligned images representations. 
As modern datasets continue to grow, the need for efficient concept-based summaries of these datasets will also grow; text-to-concept can provide such summaries easily. 

Moreoever, one can track the distribution of concept similarities for a stream of data over time. Suppose for example a model is deployed to a new setting and it begins to fail. By comparing the distribution of concept similarities in the training set to the new data, one can diagnose the distribution shifts at play. As a proof of concept, we inspect ObjectNet \cite{objectnet}, a challenging distribution shift for ImageNet models consisting of images taken within people's homes. Figure \ref{fig:objnet_vs_imgnet} shows the distribution of similarities between the vector for the concept `indoors' and aligned image representations obtained from a ResNet-50 of ImageNet and ObjectNet samples. For ObjectNet, the distribution is significantly (as determined with a Kolmogorov-Smirnov test) shifted to the right compared to ImageNet. In practice, one may maintain a bank of concepts and track similarities over their stream of data, automatically flagging concepts which experience significant shift. 

\subsection{Concept Logic for Image Retrieval}
\vspace{-0.2cm}
\begin{figure}[htb]
    \centering
    \includegraphics[width=\linewidth]{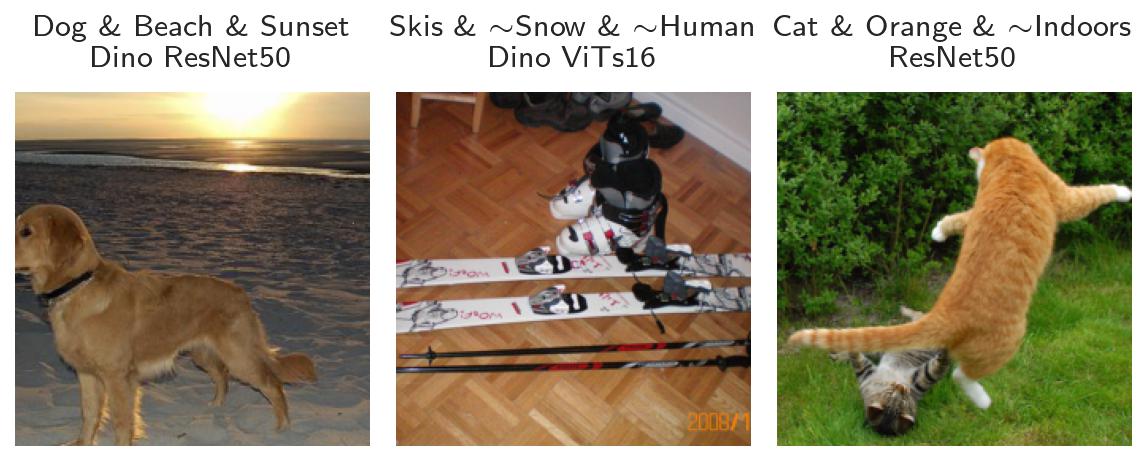}
    \caption{Images retrieved based on similarity of their representation to multiple text-to-concept vectors. Retrieved images satisfy the multiple conditions listed above each image. $\sim$ denotes `not'. Concept logic with text-to-concept enables searching over images, while allowing the use of any vision encoder to represent them.}
    \label{fig:concept_logic}
\end{figure}

Text-to-concept enables the computation of the similarity of a model's representation for an image to an arbitrary concept. Given a corpus of data and a vision encoder trained to represent said data, we can retrieve images using text, based on their similarity to text-to-concept vectors. While one may combine all keywords in a string to obtain a single composite concept vector, we observe suboptimal performance with this approach, as words receive imbalanced attention and negations are often ignored by CLIP's text encoder.

\looseness=-1
A simple alternative is to retrieve images that satisfy a set of conditions. For example, instead of searching for ``a dog on the beach at sunset'', we can separately encode the concepts `dog', `on the beach', and `at sunset'. We then filter images based on their similarity for each concept; we use thresholds based on the distribution of similarities for a given concept (i.e. at least 3 standard deviations above the mean). Analogously, we can encode negative conditions by requiring similarity to be below some threshold. Figure \ref{fig:concept_logic} demonstrates the effectiveness of our approach, retrieving rare images via concept logic (details in appendix \ref{app-sec:concept_logic}). While concept logic for image search can be done over CLIP vision embeddings, the results may be suboptimal when querying over a specific dataset for which CLIP was not finetuned, particularly compared to a vision model trained on that dataset.

\section{Concept-to-Text}
\label{sec:cct}

\looseness=-1
Text-to-concept grants insight into the representation spaces of deep models by mapping semantic notions expressed as words directly to concept vectors. However, humans still need to conjecture what concepts may be relevant before probing a representation space. We now ask, can we directly map concept activation vectors to text? We refer to this as \emph{concept-to-text}, and propose an implementation using alignment to CLIP and generative language models (Figure \ref{fig:intro-text-to-concept}).


Similar to how we that observe diverse vision models learn to store information in similar ways, allowing for cross-model alignment, we argue that language models and vision models similarly learn much of the same information, and can thus be plugged into one another with ease. Recent work supports this claim, as vision models have been stitched to generative language decoders to perform image captioning and visual question answering \cite{merullo2022linearly, magma}. Notably, \citet{clipcap} captions an image by feeding its CLIP image embedding through a finetuned version of GPT-2 \cite{GPT2}. A follow up work, ZeroCap \cite{zerocap}, similarly decodes with GPT-2 while receiving guidance from a CLIP embedding, but does so without requiring any tuning of either CLIP or GPT-2. We elect this method as it further demonstrates how existing models can work together off-the-shelf. However, other decoders could easily be put in place of ZeroCap if desired. We highlight this flexibility as it entails that concept-to-text will continue to improve as individual components are improved (e.g. GPT-2 $\rightarrow$ Chat-GPT). 
\begin{table}[t]
    \centering
    \begin{tabular}{ccc} \toprule
         Swin (S) & ResNet-50 & Dino ViTs8 \\ \midrule
          $94.48\%$ & $95.14\%$  & $92.18\%$ \\ \bottomrule
    \end{tabular}
    \caption{Percent of captions for decoded class vectors deemed relevant to images in the class by human annotators.}
    \label{tab:concept_to_text}
\end{table}

With our linear aligners, we can already map representations from off-the-shelf encoders to CLIP. Thus, with no additional training, we can perform elementary concept-to-text by simply feeding aligned features to ZeroCap. While ZeroCap expects CLIP embeddings of natural images as input, we conjecture that passing a vector encoding some semantic notion can similarly be decoded. To asses this claim, we consider the task of decoding classification head vectors. These vectors exist in the original model space, and should encode information relevant to their corresponding ImageNet class. Thus, we can quantify the effectiveness of our elementary concept-to-text method by seeing if decoded class vectors indeed describe the desired class. Specifically, we use the prompt ``Image of a '' and set the desired sequence length to 1 so that a single word is decoded per class vector. We perform a human study to answer this question, showing MTurk workers a collage of images from a given class, along with the caption obtained from decoding the class vector, and asking if this caption is relevant to the images shown. The results, shown in Table \ref{tab:concept_to_text}, show that in over $92\%$ of cases, our naive approach to concept-to-text appears effective in decoding class vectors for three diverse models. See Appendix~\ref{app-sec:cct} for additional details. 

\section{Acknowledgements}
This project was supported in part by Meta grant 23010098, NSF CAREER AWARD 1942230, HR001119S0026 (GARD), ONR YIP award N00014-22-1-2271, Army Grant No. W911NF2120076 and the NSF award CCF2212458.




\bibliography{references}
\bibliographystyle{icml2023}

\newpage
\appendix
\onecolumn
\section{Cross-Model Alignment}
\label{app-sec:alignment}


\begin{figure*}[tbp]
\centering
    \begin{subfigure}{0.48\linewidth}
        \includegraphics[width=1\linewidth]{figures/general_alignment/RA/agg-RA.pdf}
        \caption{The average of retained accuracy.}
    \end{subfigure} \hfill
    \begin{subfigure}{0.48\linewidth}
        \includegraphics[width=1\linewidth]{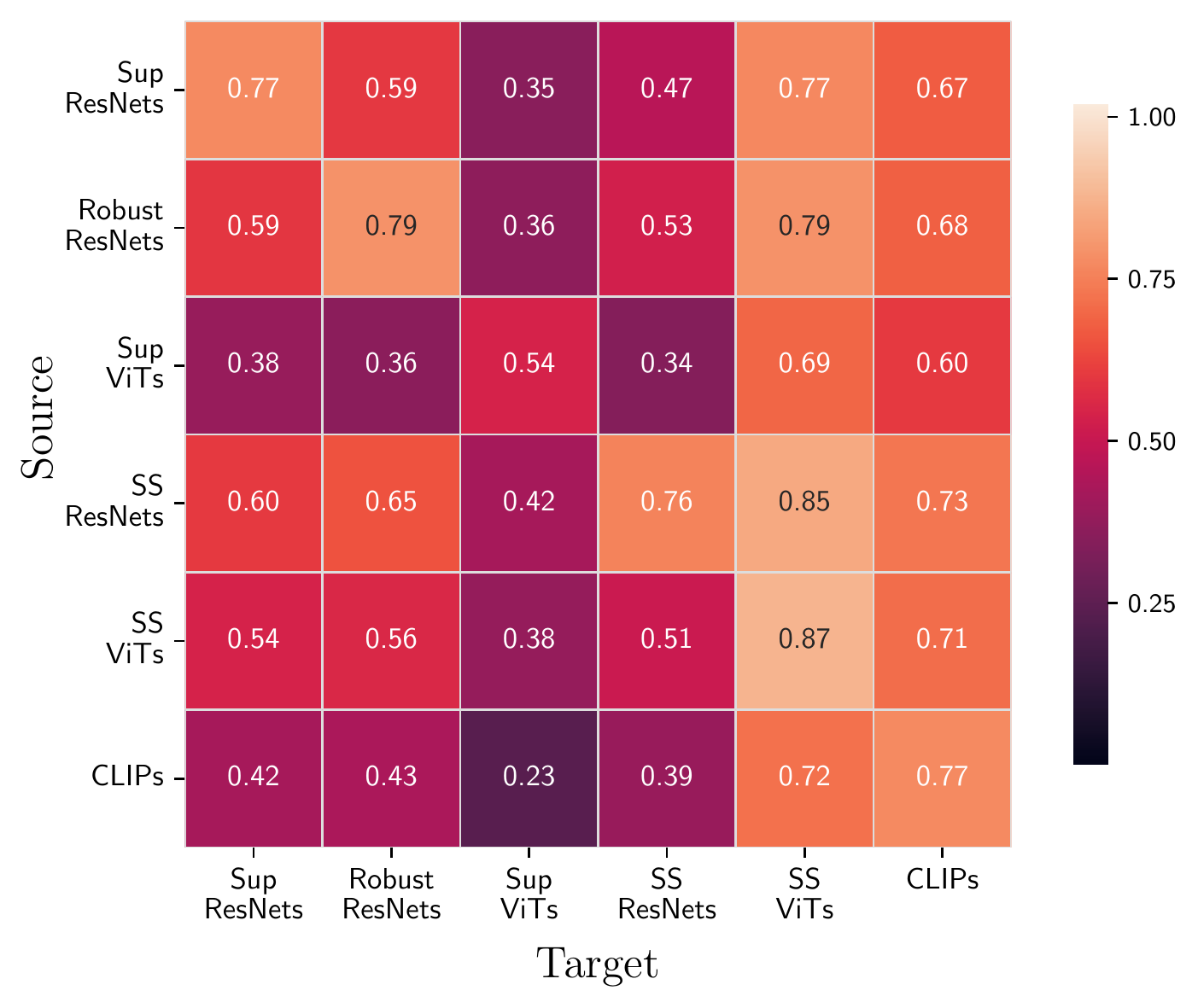}
        \caption{The average of $\Rsquare$.}
    \end{subfigure} \hfill
    \caption{
        \textbf{(Left)} shows the average of retained accuracy.
        More precisely, 
        the value in row $r$ and column $c$ is the average of retained accuracy when doing alignment from representation space of model $s$ to that of model $t$
        where $s$ is a model in group $r$ and $t$ is a model of group $c$. 
        \textbf{(Right)} shows the average of $R^2$, i.e., same as above, value in row $r$ and column $c$ is the average of $R^2$ in linear alignment
        from models in group $r$ to models of group $c$. Note that ``Sup" stands for Supervised while ``SS" stands for Self-Supervised training procedure.
        Note that all models are pretrained on ImagenNet-1K except CLIPs. More details on models used here can be found in Section~\ref{app-sec:models}.    
    }
    \label{fig:gen-alignment}
\end{figure*}
  
\begin{figure*}[tbp]
\centering
    \includegraphics[width=0.8\linewidth]{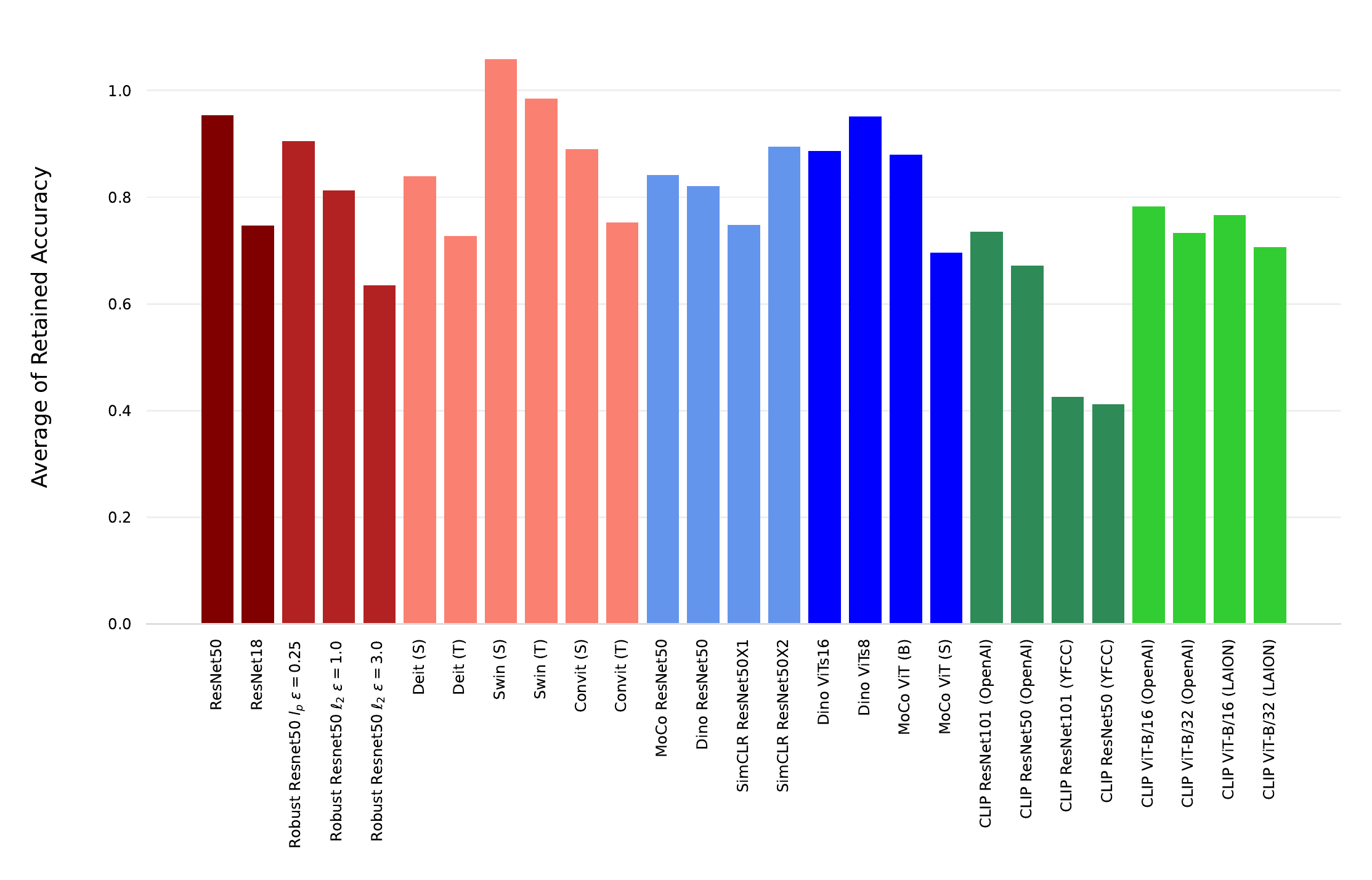}
    \caption{
    For each model $s$, average of retained accuracy when doing alignment from model $s$ to all other models is reported.
    }
    \label{fig:fixed-source-all}
\end{figure*}


In this section, we conduct an extensive set of experiments
where we evaluate linear alignment between many pairs of models (see Section~\ref{app-sec:models} for more details on models).
We also consider cases where two models are significantly different, i.e.,
they may have different architectures, or they are trained with different procedures (supervised learning, self-supervised learning, etc).

we formally define \emph{aligned accuracy}, i.e., the accuracy of alignment on $\Dtest$, 
and \emph{retained accuracy}, i.e., the ratio of aligned accuracy to the target model accuracy,
when linear alignment is done from the representation space of $f_s$ to that of $f_t$.
Note that $W, b$ are the solutions of optimization problem provided in Section~\ref{sec:alignment}.

\begin{align*}
\text{aligned accuracy} :=
    \frac{1}{|\Dtest|} \sum_{(x, y) \in \Dtest} \ind{g_{t}\left(h_{W, b}(x)\right) = y}
\\
\text{retained accuracy} :=
    \frac{
        \text{aligned accuracy}
    }{
        \frac{1}{|\Dtest|} \sum_{(x, y) \in \Dtest} \ind{g_{t}\left(f_t(x)\right) = y}
    }
.
\end{align*}

As seen in Figure~\ref{fig:gen-alignment}, linear alignment generally works well in the sense of both $\Rsquare$ and retained accuracy.
In many cases, we see $\Rsquare$ score above $0.6$,
which indicates that a significant portion of variance is captured in the linear regression.
We also see better values for $\Rsquare$ on the diagonal of right heatmap in Figure~\ref{fig:gen-alignment} as
representation spaces of models with the same architecture or training procedure are more linearly transformable.
Furthermore, models are surprisingly capable of retaining the accuracy of the target models
when a linear transformation is done on their representation spaces.
Some models are more capable of retaining accuracy, e.g., as seen in Figure~\ref{fig:gen-alignment}, {\emph Supervised Vision Transformers}
get even better accuracy when their feature spaces are \textbf{linearly transformed} into other models' feature spaces and
the \textbf{classification heads of the other models} are applied.
Figure~\ref{fig:fixed-source-all} shows the average of retained accuracy
for each individual source model $s$ along
the average of retained accuracy for each group of source models.

This is due to the fact that richer and more informative representation spaces make classification easier.
Note that in our experiments, all models except CLIPs are trained on ImageNet-1K dataset \cite{deng2009imagenet}
but we evaluate all of them with ImageNet.
Note that this dataset is remarkably hard and big, which 
further supports the idea that linear alignment is possible and according to Figure~\ref{fig:gen-alignment},
different models are actually learning linearly transformable concepts.

\begin{figure*}[tbp]
\centering
    \begin{minipage}{0.48\linewidth}
    {\includegraphics[width=\linewidth]{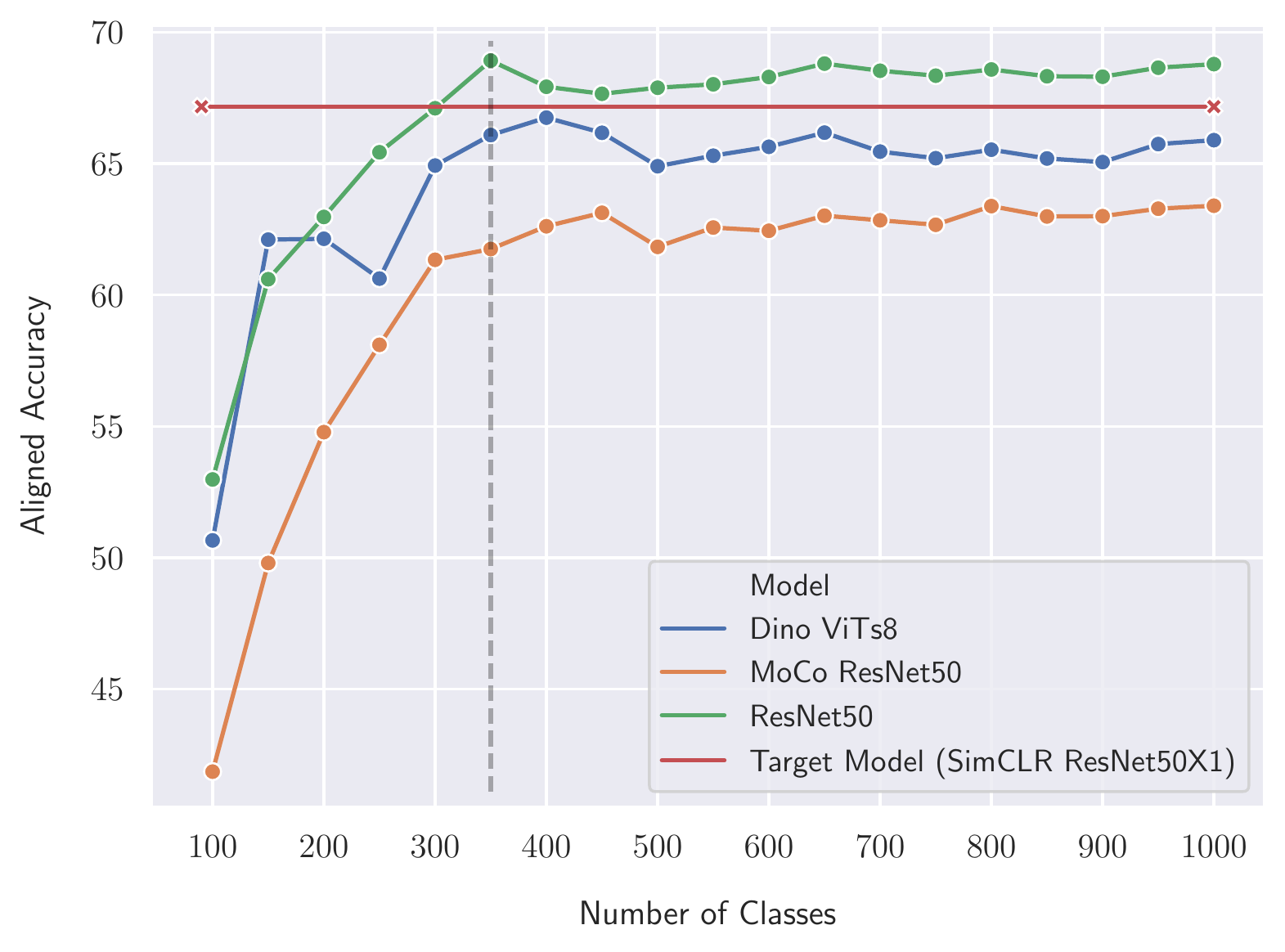}} 
    \end{minipage} \hfill
    \begin{minipage}{0.48\linewidth}
    {\includegraphics[width=\linewidth]{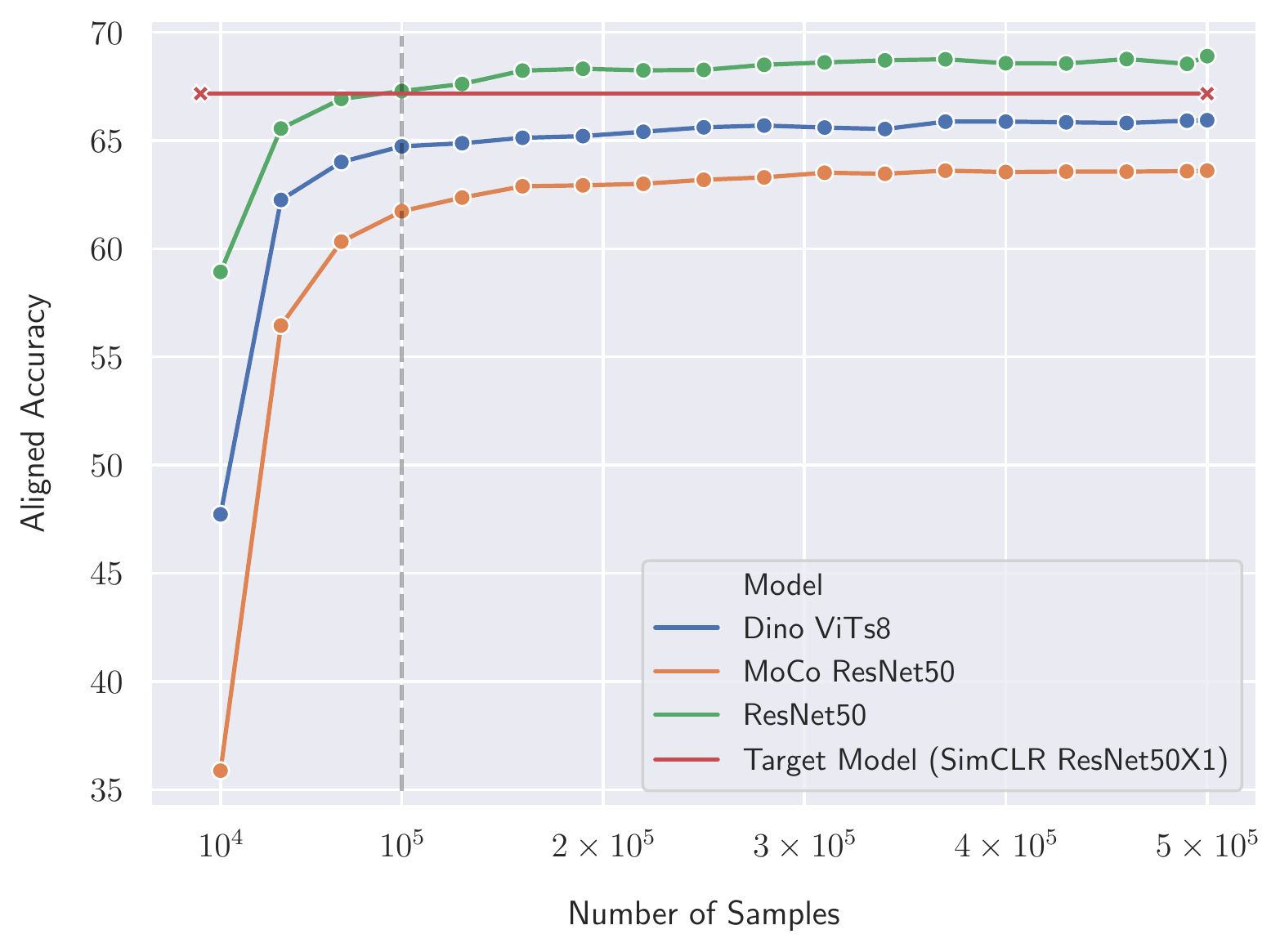}}
    \end{minipage} \hfill
    \caption{
    \textbf{(Left)} 
    shows the aligned accuracy when
    linear transformation is only optimized on images with particular labels.
    We randomly select labels and increase the number of labels(classes) to see how retained accuracy changes.
    while \textbf{(Right)} 
    shows the aligned accuracy when linear alignment is solved on a random subset of images.
    Alignment is done from three different models to SimCLR ResNet50X1.
    We observe that all training images are not necessary to have a~reliable alignment.
    In other words, aligned accuracy can reach to its maximum by only considering small portion of images or classes.
    }
    \label{fig:sample-eff-gen-alignment}
\end{figure*}

\subsection{Optimizing Linear Transformation}
\label{app-sec:alignment-opt}
Setting $\Dtrain := \{x_i\}_{i=1}^N$ and $\Dtest := \{x_i\}_{i=1}^{N'}$, we note that the optimization problem to obtain $W$ and $b$ is
\begin{align}
    W, b = \argmin_{W, b} \frac{1}{N} \sum_{i=1}^{N} \left(W^T f_{s}(\xtrain_i) + b - f_{t}(\xtrain_i)\right)^2    .
\end{align}
\label{eq:gen-alignment2}

With a proper set of hyperparameters 
around 6 epochs are enough to converge to the optimal solution.
However, 
re-scaling representation spaces of models so that the variance of elements in the space becomes constant,
is crucial (See Section~\ref{app-sec:opt}).
This is due to the fact that some models embed inputs into very low variance spaces,
which degrades the performance of linear alignment due to precision in computations.

Additionally, we take into account the optimization problem given in \eqref{eq:gen-alignment2} and
consider the effect of the number of images that we involve in optimizing \eqref{eq:gen-alignment2}.
We observe that using only a random subset of the training set of ImageNet
is sufficient to find $W$ and $b$, as seen in Figure~\ref{fig:sample-eff-gen-alignment}, $1/5$ of ImageNet training samples
is roughly enough to retrieve the target model accuracy.
if we use only images of some particular classes to optimize \eqref{eq:gen-alignment2}, 
we can retrieve the target accuracy by just using around $1/3$ of ImageNet classes.

\subsection{Alternate Objectives for Alignment to CLIP}

We now present alternate objectives to enable alignment, specifically to vision language models like CLIP. The method we present optimizes a linear layer aligner for the regression task of predicting features for one model given features from another. An alternate approach is to optimize the aligner \emph{directly for classification}. That is, we use a cross entropy loss $\ell$, obtaining logits for a sample $x$ by passing features $f(x)$ from a fixed encoder $f$ through a trainable linear aligner with parameters $W,b$ before finally having their cos-sine similarity taken with a set of text embeddings for each class obtained via CLIP’s text encoder. To recap, we solve $\min_{W,b} \sum_{x,y\in D} \ell(\text{sims}(x), y)$ for a dataset 
$D$ of labeled images (x,y), where $\text{sims}(x)_i = \text{cossim}(W^Tf(x) + b, t_i)$, where $t_i$ is the CLIP text embedding of the name of class $y_i$. Essentially, we fix a classification head in CLIP space as the text embeddings of the class names for a task. Then, we optimize a layer stitching a given image encoder to this classification head. 

The requirement of labeled data is one drawback of this baseline compared to our method of aligning representations. Secondly, this training does not necessarily entail that the aligner truly learns to map to the target representation space, as any information irrelevant to the selected class vectors may be lost. Thus, while zero-shot performance on tasks similar to the one optimized for will likely exceed our method, it may come at the cost of significantly reduced performance on unrelated tasks. Indeed, we find this to be the case: when optimizing the aligner for ImageNet classification, we observe the alternate method to perform roughly $11\%$ worse on zero-shot classification tasks unrelated to ImageNet, though performance on ImageNet-like tasks is roughly $5\%$ better (see table \ref{tab:compare_to_alternate}). 

\begin{table}[]
    \centering
    \begin{tabular}{c|c|cc} \toprule
         & Overall & ResNets	& ViTs \\ \midrule
         Related to ImageNet	& -4.99\% & -1.50\%	&-8.47\% \\
         Not Related to ImageNet	& 11.21\% & 14.70\%	& 7.72\% \\ \bottomrule
    \end{tabular}
    \caption{Average gain in zero-shot accuracy of the original aligner optimization (i.e. regression on image features) over the alternate (i.e. cross entropy loss using fixed classification head). Two ResNet50s (standard, MoCo) and two ViTs (DeiT, MoCo) considered. Datasets denoted as “Related to ImageNet” are: ImageNet9 and all BREEDS datasets (direct coarse grained categorizations of ImageNet images), and CIFAR10, STL10, and Fashion MNIST (ImageNet classes/coarse categories in OOD data). Datasets denoted as “Not Related to ImageNet” are: CelebA hair classification, DTD, Colors, Shapes, SVHN.}
    \label{tab:compare_to_alternate}
\end{table}

We note that the ViTs we consider benefit much more from the alternate method than ResNets. We conjecture this is due to the lower dimensionality of ViT feature space relative to CLIP, which makes our linear regression task under-determined and thus challenging. In contrast, the alternate method provides a less stringent optimization task, since the aligner needs only to map representations of samples from different classes to separable clusters in CLIP space. This makes the training easier and more successful for ViTs, though it comes at two key limitations: the baseline requires labeled data, and the alignment is less reliable for concepts that are not directly related to the classification task the aligner is optimized for.

Finally, it may be possible to solve for alignment in a completely data-free manner. Namely, one may seek to solve a matrix equation to re-express an existing classification head as the product of a learnable aligner and a fixed classification head in CLIP space (e.g. obtained by embedding the names of the classes for the task). In addition to being data-free, this method potentially would result in faithful preservation of the original model's behavior, while still unlocking all of the interpretability benefits of text-to-concept. Given the importance of faithfulness in interpretability, and the flexibility of not requiring data, we believe solving alignment in this manner may be a promising line of future investigation. 

\section{PC Alignment}
\label{app-sec:pca}
\begin{figure}[tbp]
\centering
\begin{minipage}{.41\textwidth}
  \centering
  \includegraphics[width=.9\linewidth]{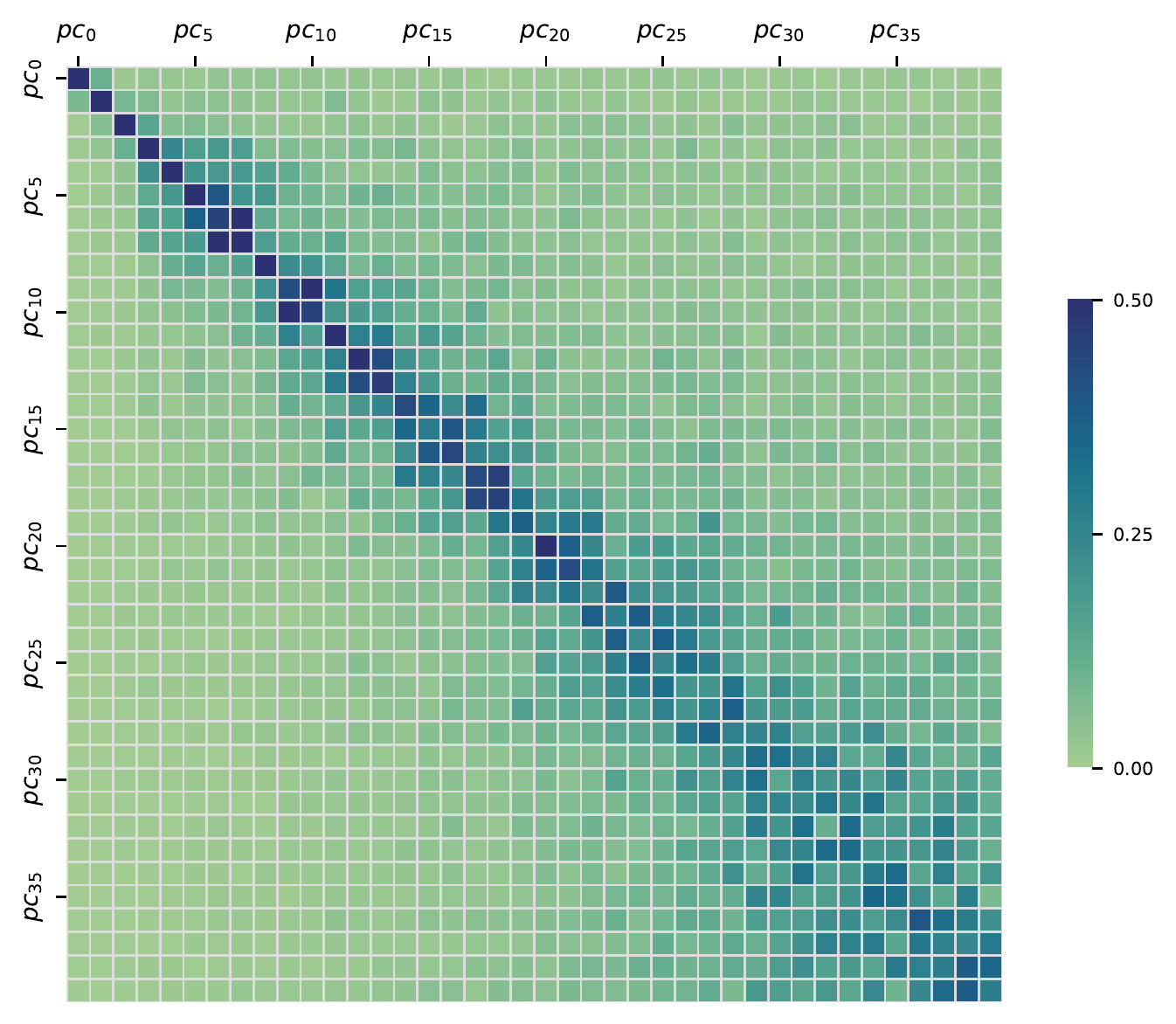}
  \captionof{figure}{
    Shows the heatmap of the average of matrix $W^T$, when doing alignment between every pair of $4$ CLIP models pre-trained on OpenAI dataset. Two of these $4$ models use ResNet architecture while two others use Vision Transformers. We observer that the matrix is almost diagonal, which implies that there is a 1-1 relation between top principal components in CLIP models.
  }
  \label{fig:pca-alignment1}
\end{minipage} \hspace{10pt}
\begin{minipage}{.49\textwidth}
  \centering
  \includegraphics[width=.9\linewidth]{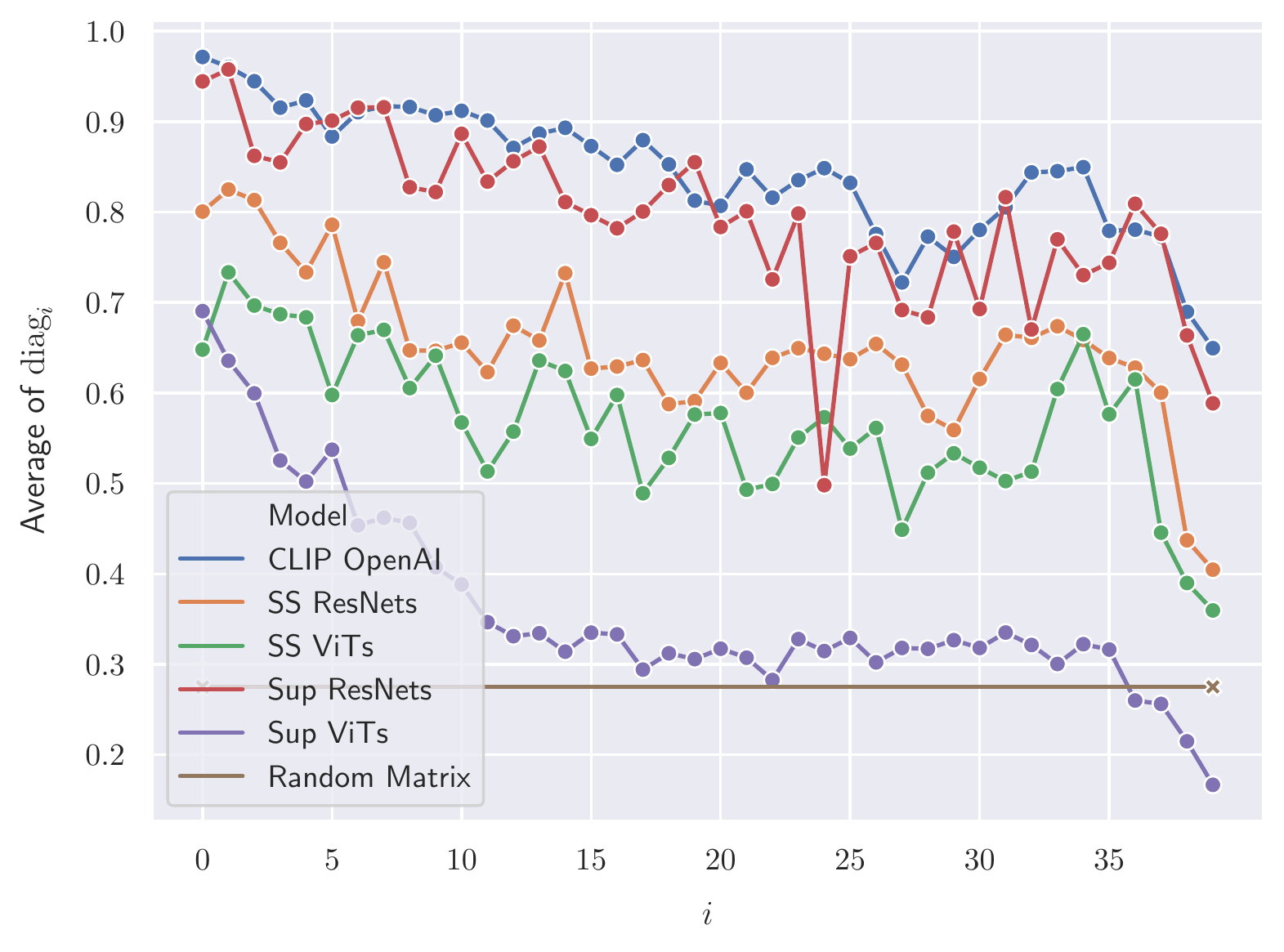}
  \captionof{figure}{
    Shows the plot of the average of $\diag_i$ when we do linear alignment between all pairs of models within each group of them.
    Generally, we see larger values for $\diag_i$ when $i$ is smaller. This implies that 1-1 correspondence holds more for top principal components. 
    High values of $\diag_i$ show that the geometry of principal component space of top components
    within CLIP models, supervised ResNets, and self-supervised ResNets are approximately same. For more details regarding models, see Section~\ref{app-sec:models}.
  }
  \label{fig:pca-alignment2}
\end{minipage}
\end{figure}


In this section, we extend linear alignment to the space of principal components. 
We know that the effective dimension of representation spaces of models is relatively low.
This is mostly due to the existence of redundant information in these spaces so that
many of the features can be approximated with a linear combination of others
or the existence of some noise features.
As a result, it is reasonable to consider the representation spaces in a more abstract manner.
To do so, we use \emph{Principal Component Analysis} (PCA) \cite{wold1987principal}.
Let $\pc_0, \pc_1, ..., \pc_{d-1}$ be the principal components of
the representation space of model $m$, i.e., $\{f_m(\xtrain_i)\}_{i=1}^{N}$,
in decreasing order of their corresponding eigenvalues
\footnote{Note that before doing PCA, we centralize our points in representation space such that mean of points become $\vec{0}$.} 
.
We take top $k$ principal components and project representations into the space of these components.
For each point $f_m(x)$, we get a vector $q_m(x) \in \mathbb{R}^{k}$ where
\begin{align*}
    q_m(x) := \left(
        f_m(x) \cdot \pc_i 
    \right)_{i=0}^{k-1}
    .
\end{align*}

Now, for each input image $x$,
we have a low-dimension embedding which due to the properties of PCA,
retains a~significant portion of the information of the original representation $f_m(x)$.
Like Section~\ref{sec:alignment},
we apply linear alignment on \textbf{principal components space} of model $s$ to get that of model $t$. 
Formally, we define linear transformation $h(z)$ as
\begin{align*}
    h(z) := W^T z + b
    .
\end{align*}

To find best $W$ and $b$,
we follow \eqref{eq:gen-alignment} but replace $f$ with $q$.
Note that PC alignment,
matrix $W$ and vector $b$ have significantly lower dimensions, i.e.,
$W \in \mathbb{R} ^ {k\times k}$ and $b \in \mathbb{R} ^ k$.

Interestingly, we see that between many pairs of models, even though there is a significant difference in architecture, 
there approximately exists a 1-1 correspondence between principal components.
Indeed, $i$-th element of $q_t(x)$ can be approximated by $i$-th element of $q_s(x)$.
This implies that surprisingly, (1) \textbf{top principal components represent the same abstract knowledge} even in different models, 
and (2) the\textbf{ order of these top components is preserved} among models. 
A visualization of matrix $W^T$ where each row is normalized is depicted in Figure~\ref{fig:pca-alignment1}. 
We measure the observation of 1-1 correspondence in a quantitative manner.
Note that according to linear alignment,
\begin{align*}
    q_t(x)_i \approx \sum_{j=1}^{k} W^T_{i, j} q_s(x)_j
    ,
\end{align*}
where $q_s(x)_j$ denotes the $j$-th element of $q_s(x)$ and $q_t(x)_i$ denotes the $i$-th element of $q_t(x)$.
We normalize each row in $W^T$ and then measure how much elements close to the diagonal
contribute in the approximation of $i$-th element, i.e.,
for each $i \in \{0, 1, ..., k-1\}$, we define $\diag_i$ as
\begin{align*}
    \diag_i := \sum_{j=\max\left(0, i-p\right)}^{\min\left(k - 1, i+p\right)} \left(W^T_{i, j}\right) ^ 2
    ,
\end{align*}
where $p = 5$ in our experiments. Also we take $k = 40$ top principal components.
Figure~\ref{fig:pca-alignment2}, shows the average value of $\left(\diag_i\right)_{i=0}^{k-1}$
when doing alignment between every pair of models in different groups.
We consider $5$ groups of models (see Section~\ref{app-sec:models} for details about each group)
with different architecture or training procedure.
As observed in Figure~\ref{fig:pca-alignment2},
1-1 correspondence holds more within CLIP models, supervised, and self-supervised ResNets while it doesn't hold within supervised vision transformers.

\section{Models}
\label{app-sec:models}
In this paper we have considered several different models in different categories \cite{Radford2021LearningTV}, \cite{moco}, \cite{dino}, \cite{he2016deep}, \cite{salman2020adversarially}, \cite{Radford2021LearningTV}.
All these models except CLIP models are trained on ImageNet-1K \cite{deng2009imagenet}.
For almost all of these models, pretraiend weights are obtained from \href{https://github.com/rwightman/pytorch-image-models}{timm} library \cite{timm} and \cite{Ilharco_Open_Clip_2021}.
Our models are categorized in following groups.
\begin{itemize}
    \item \textbf{Supervised ResNets} include ResNet50, and ResNet18.
    \item \textbf{Robust ResNets} include Robust Resnet50 $\ell_2,\epsilon=0.25$, Robust Resnet50 $\ell_2,\epsilon=1.0$, and Robust Resnet50 $\ell_2,\epsilon=3.0$.
    \item \textbf{Supervised Vision Transormers} include Swin, Deit, and Convit models.
    \begin{itemize}
        \item Swin with patch size of 4 and window size of 7 includes Swin Small (S) and Swin Tiny (T).
        \item Deit with patch size of 16 includes Deit Small (S) and Deit Tiny (T).
        \item Convit includes Convit Small (S) and Convit Tiny (T).
    \end{itemize}
    \item \textbf{Self-Supervised ResNets} include MoCo ResNet50, Dino ResNet50, SimCLR ResNet50X1, and SimCLR ResNet50X2.
    \item \textbf{Self-Supervised Vision Transformers} include MoCo~ViT~base~(B), MoCo~ViT~small~(S), Dino~ViTs~16, and Dino~ViTs~8, .
    \item \textbf{CLIP} include
    \begin{itemize}
        \item CLIP ResNet101, CLIP ResNet50, CLIP ViT-B/16, and CLIP ViT-B/32 trained on OpenAI dataset.
        \item CLIP ResNet101 and CLIP ResNet50 trained on YFCC \cite{thomee2016yfcc100m}.
        \item CLIP CLIP ViT-B/16 and CLIP CLIP ViT-B/32 trained on LAION \cite{schuhmann2022laion}.
    \end{itemize}
\end{itemize}

\section{Optimizing Linear Alignment}
\label{app-sec:opt}
In terms of optimizing \ref{eq:gen-alignment2},
we use SGD optimizer and learning rate scheduler (implemented in Torch \cite{pytorch}) with following hyperparameters:


\texttt{optimizer = optim.SGD(lr=0.01, momentum=0.9, weight\_decay=5e-4)}

\texttt{scheduler = torch.optim.lr\_scheduler.CosineAnnealingLR(T\_max=200)}

We run optimization for $6$ epochs.
Note that before optimizing we re-scale representation spaces of models such that variance of elements
in matrix $\left(f_s(\xtrain_1), f_s(\xtrain_2), ..., f_s(\xtrain_N)\right)$ becomes $4.5$.

\section{Prompts for Text-to-concept}
\label{app-sec:ttc}

\begin{figure*}
    \centering
    \subfloat{\includegraphics[width=\textwidth]{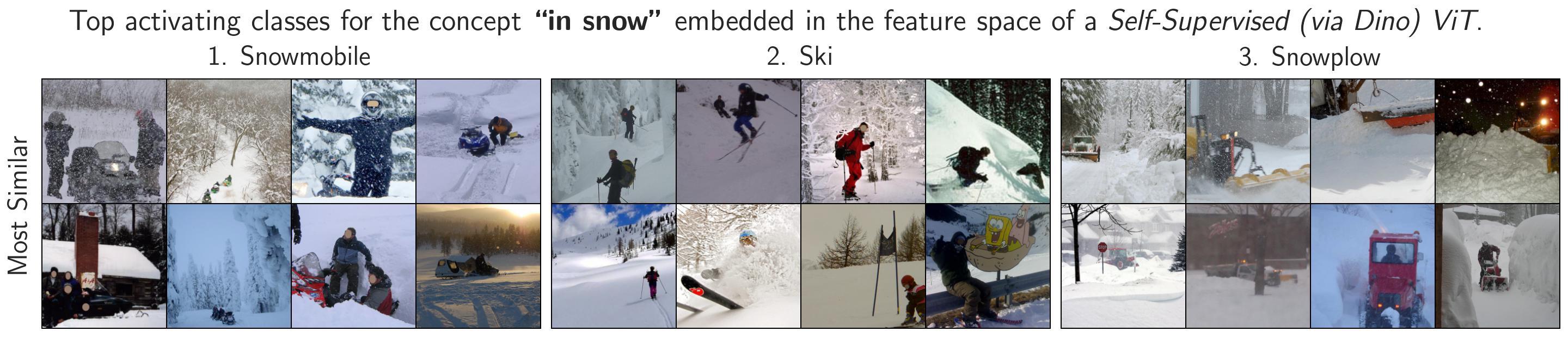}} \\ 
    \vspace{-0.5cm}
    \subfloat{\includegraphics[width=\textwidth]{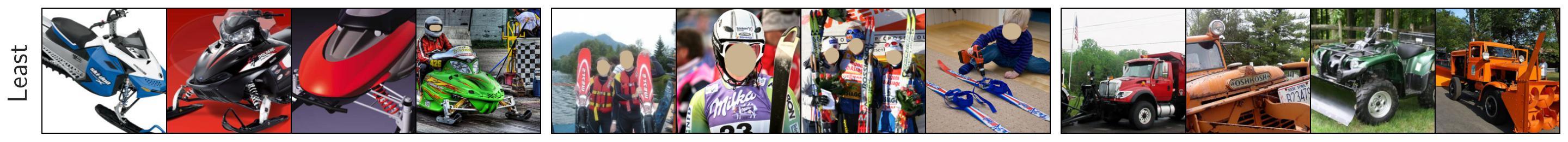}}\vspace{-0.1cm}
    \caption{Additional qualitative validation of text-to-concept. Observe that the samples least similar to the concept vector for ``in snow'' break a common spurious correlation for their classes. Text-to-concept may then be used to identify challenging natural images within datasets, towards mitigating spurious correlation dependencies.} 
    \label{fig:eg_in_snow}
\end{figure*}

When not otherwise specified, we use the default templates introduced in the original CLIP paper for ImageNet zero-shot classification. They are as follows: \textit{`itap of a \{\}', `a bad photo of the \{\}', `a origami \{\}', `a photo of the large \{\}', `a \{\} in a video game', `art of the \{\}', `a photo of the small \{\}'}. 

For Figure \ref{fig:eg_in_a_tree} and Figure \ref{fig:eg_in_snow}, we append ``in a tree'' and ``in snow'' to the above templates, and also replace the '\{\}' with names for all ImageNet classes. The final concept vector is then an average of $\textsc{number of templates} \times \textsc{number of clases}$ vectors. We do this because these correspond to contexts, which should be object agnostic. Similar results are obtained without refinement (i.e. replacing $\{\}$ with `object'). We note that obtaining embedding text to CLIP's space is very quick, only taking seconds to encode a batch of thousands of short phrases.

\begin{table}[htb]
    \centering
    \begin{tabular}{c|c|c|c} \toprule
        Dataset & Example Classes & Prompt & Citation \\ \midrule
        \multicolumn{4}{c}{Coarse Grained Concepts In Distribution} \\ \midrule
IN9 & dog, bird, wheeled vehicle & a photo of \{\} &  \citet{in9} \\
Living17 & salamander, turtle, lizard & a photo of a \{\} &  \citet{breed} \\
Nonliving26 & bag, ball, boat & a photo of a \{\} &  \citet{breed} \\
Entity13 & garment, bird, reptile & a photo of a \{\} &  \citet{breed} \\
Entity30 & serpentes, passerine, saurian & a photo of a \{\} &  \citet{breed} \\ \midrule
\multicolumn{4}{c}{Coarse Grained Concepts Out of Distribution} \\ \midrule
CIFAR10 & airplane, automobile, bird & a pixelated photo of a \{\} &  \citet{cifar10} \\
STL10 & airplane, bird, car & a photo of a \{\} &  \citet{stl10} \\
Fashion
MNIST & T-shirt/top, Trouser, Pullover & a black and white photo of \{\} &  \citet{Xiao2017FashionMNISTAN}  \\
CelebA Hair & brown hair, blonde hair & a headshot of a person with \{\} &  \citet{liu2015faceattributes} \\ \midrule
\multicolumn{4}{c}{Character Recognition}\\ \midrule
SVHN & zero, one, two & a photo of the digit ``\{\}" on a building &  \citet{svhn} \\
MNIST & zero, one, two & a photo of the digit ``\{\}" &  \citet{mnist} \\ \midrule
\multicolumn{4}{c}{Primitive Concepts}\\ \midrule
Textures & banded, blotchy, braided & a photo of something with \{\} texture &   \citet{dtd} \\ 
Color & black, blue, brown & a swatch of the color \{\} &  - \\
Shape & circle, octagon, square & a diagram of the shape \{\} &  \citet{shapes} \\ \bottomrule
    \end{tabular}
    \caption{List of datasets studied in Zero-Shot classification experiments (Section \ref{sec:zero_shot}), along with example classes and the specific prompt used. Note that we use an internal simple dataset for \emph{Color}.}
    \label{tab:zero_shot_details}
\end{table}

\section{Zero-shot Classification}
\label{app-sec:zero_shot}

We carry out zero-shot experiments over many datasets. We use slightly different prompts for each task, though we stress that we did not optimize prompt engineering to obtain better results. All evaluated models use the same prompts. Table \ref{tab:zero_shot_details} shows details for prompts used, as well as example classes for each dataset, to give an idea as to what kind of text is used to generate concept vectors. We refer readers to the original sources for more details on the datasets studied.  

For color recognition, we construct a simple dataset that consists of one sample per the following classes: \emph{black, blue, brown, gray, green, orange, pink, purple, red, white, yellow}. The sample in each class is a monocolor patch, with every pixel set to have the color given by the class name. Also, for shape recognition, we use a subset of the shapes in the original dataset. Specifically, we include the following shapes: \emph{circle, octagon, square, star, triangle}.

\section{Concept Bottleneck Models}
\label{app-sec:cbn}

\begin{figure*}
    \centering
    \includegraphics[width=\linewidth]{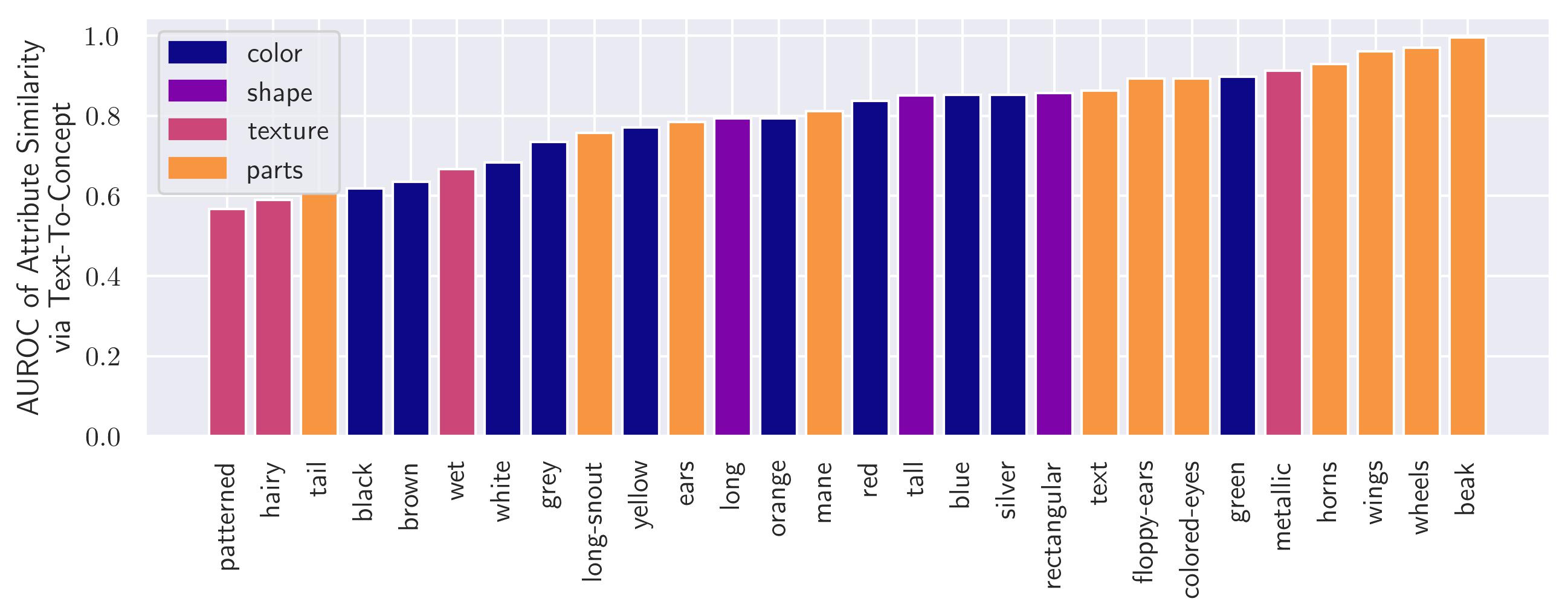}
    \caption{Quality of using similarity to text-to-concept vectors for predicting RIVAL10 attributes. AUROC shown per attribute. Attributes corresponding to parts are predicting more reliably. Over $70\%$ of attributes achieve an AUROC of at least $0.75$.}
    \label{fig:attr_preds}
\end{figure*}

\begin{figure*}
    \centering
    \begin{minipage}{0.32\textwidth}
    \centering
    \includegraphics[width=0.99\textwidth]{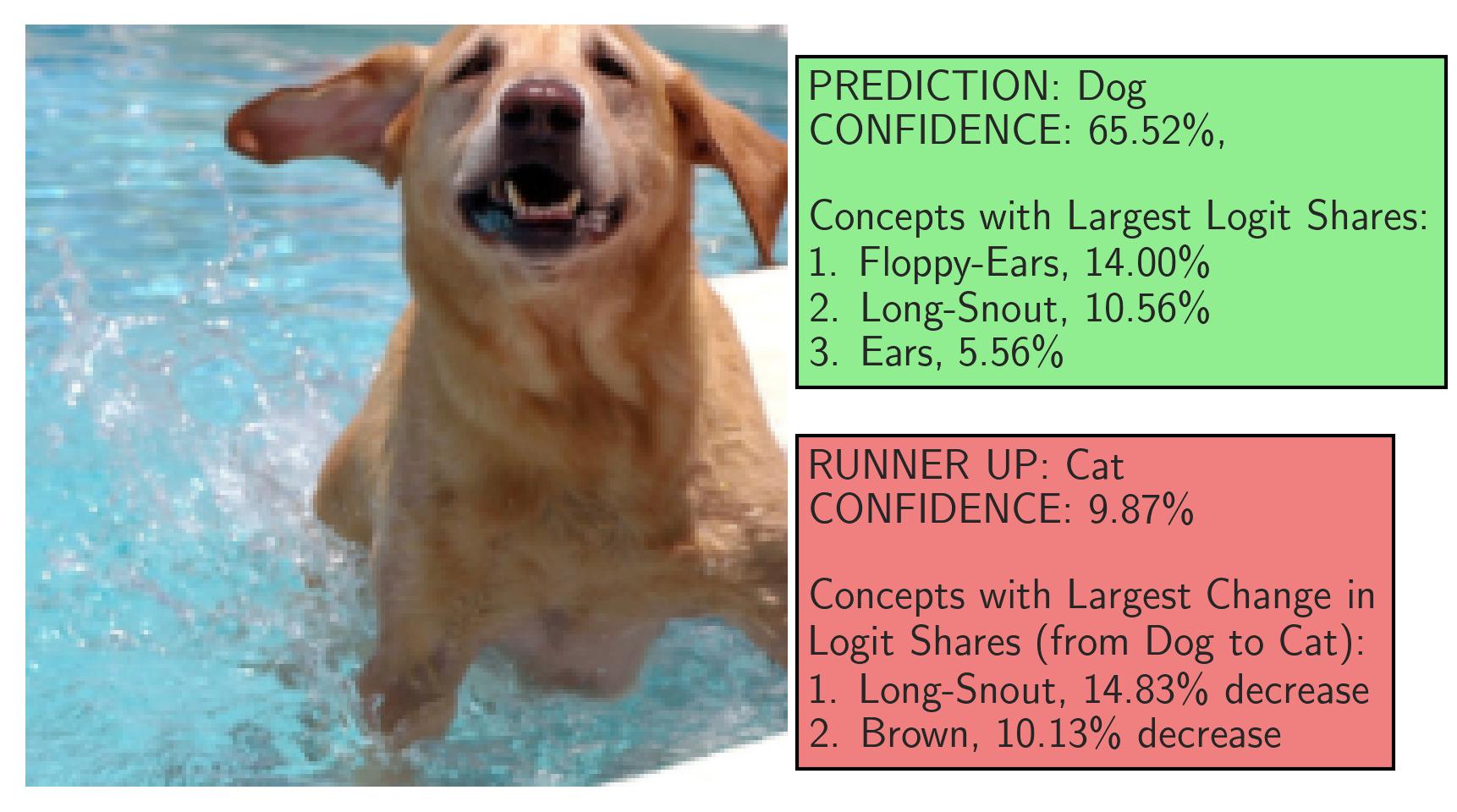}
    \end{minipage} %
    \begin{minipage}{0.32\textwidth}
    \centering
    \includegraphics[width=0.99\textwidth]{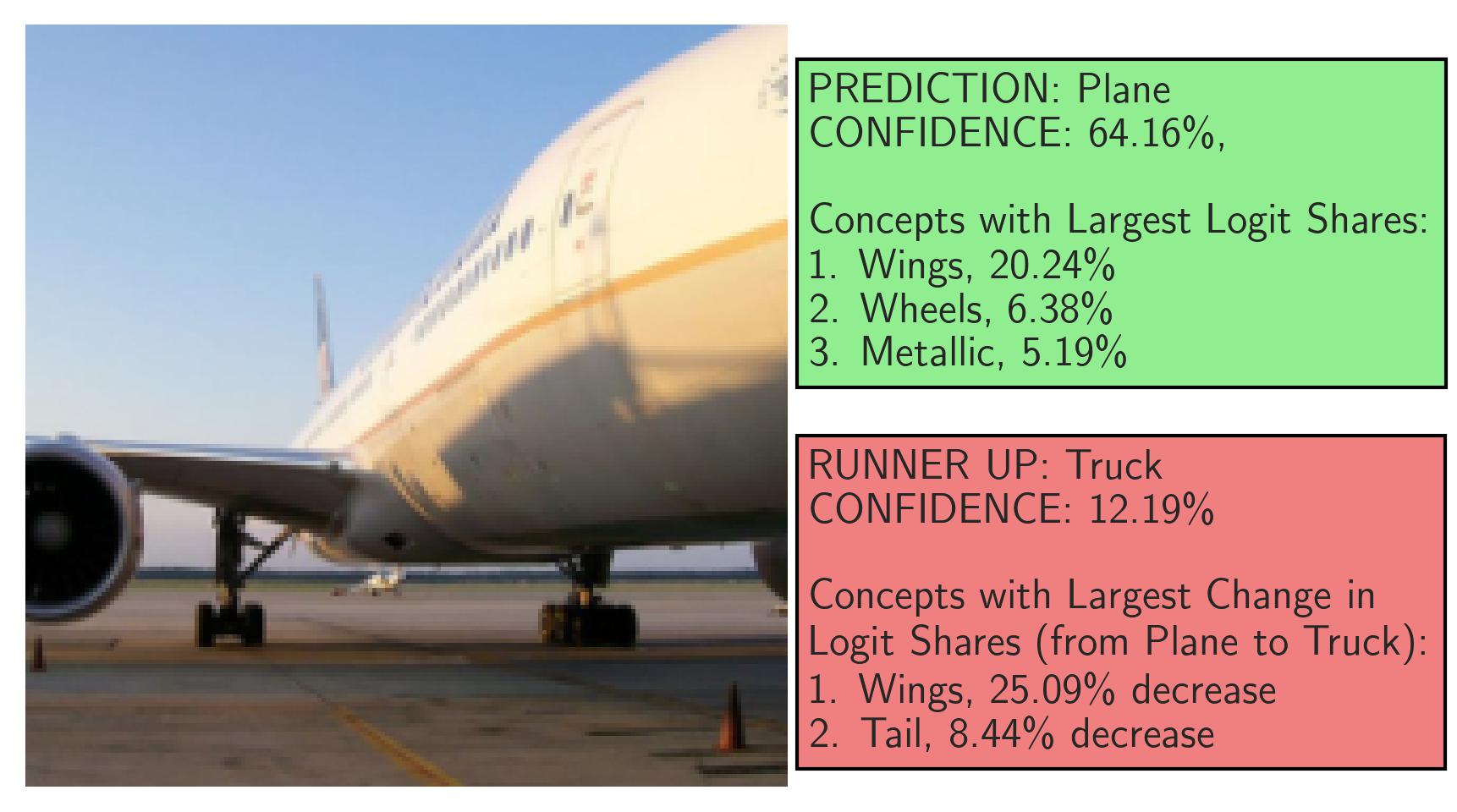}
    \end{minipage}
    \begin{minipage}{0.32\textwidth}
    \centering
    \includegraphics[width=0.99\textwidth]{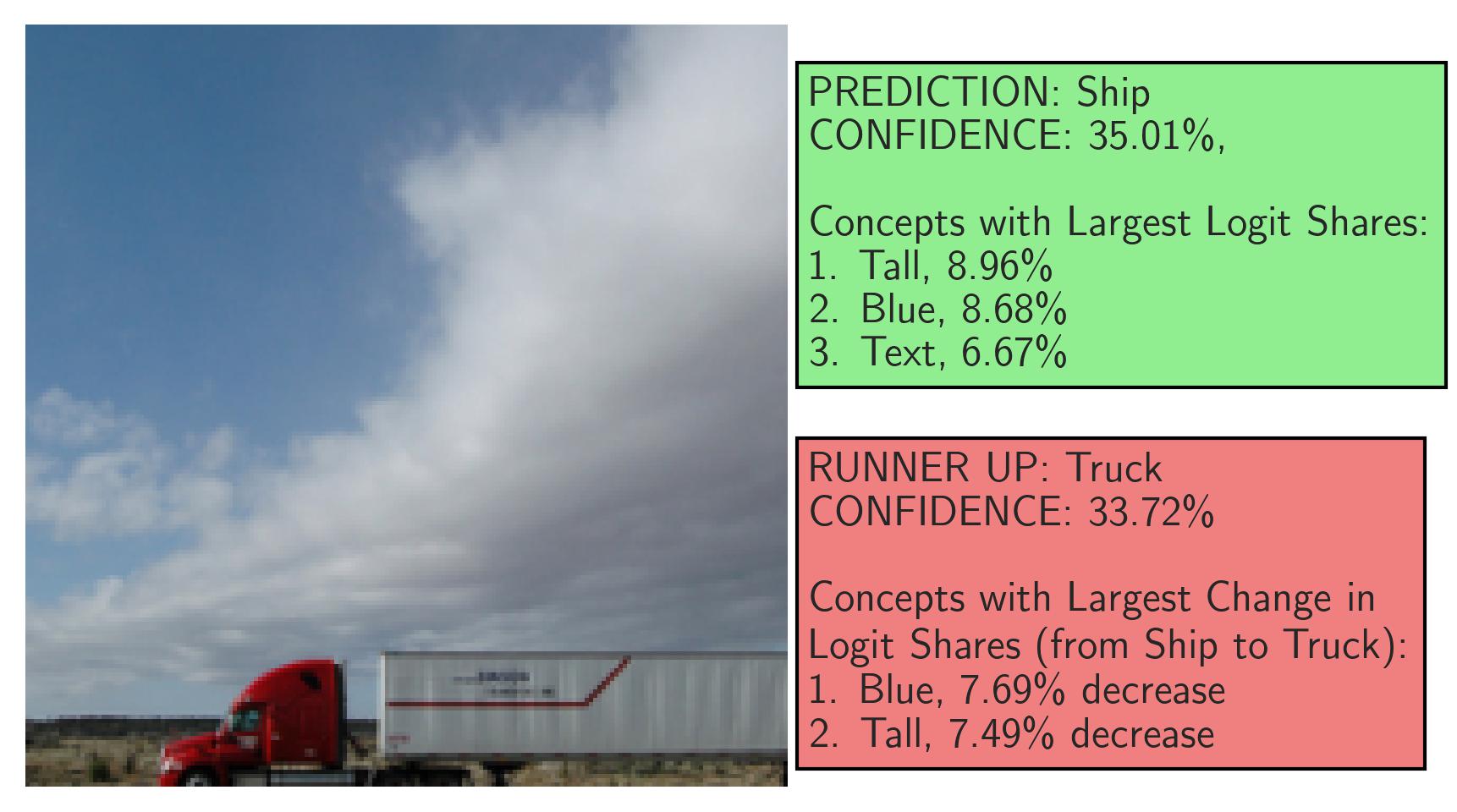}
    \end{minipage}
    \caption{Extra examples of inference using the concept-bottleneck model (enabling direct measurement of each concept's contribution to a class logit, as shown) built atop a fixed vision encoder and text-to-concept. We include a misclassification in the rightmost panel.}
    \label{fig:cbn_extra_egs}
\end{figure*}

RIVAL10 \cite{rival10} is a ten class classification problem operating on ImageNet images. Each image is also annotated with $28$ attributes, though we only use these to evaluate our associated text-to-concept vectors. It is analagous to the classes in CIFAR10. Training the CBM is quite simple: 
\begin{enumerate}
    \item We obtain features for each image using a pretrained ResNet50.
    \item We obtain vectors corresponding to each RIVAL10 attribute. We use standard templates with no class averaging (simplest mode, no prompt engineering).
    \item We use our trained aligner from ResNet50 to CLIP ViT-B/16 (section \ref{sec:alignment}) to align features to CLIP space. Recall this is just feeding saved features to a linear layer.
    \item We compute and save the cosine similarity of each aligned image representation to the attribute concept vectors obtained in step 2. Again, this amounts to normalizing two matrices before multiplying them together.
    \item We train a linear layer mapping the similarities (obtained in previous step) of aligned representations and attribute concept vectors to RIVAL10 class labels. We train this linear layer for 40 epochs with SGD. 
\end{enumerate}

One can see that the conversion of an existing encoder to the CBM for the desired task is simple and requires minimal training (we only train a linear layer for the aligner and one for the classification head). We achieve $93.8\%$ accuracy, and AUROCs for each attribute prediction as shown in figure \ref{fig:attr_preds}. Note that text-to-concept vectors for color attributes seemed to be the least reliably, while vectors for part attributes are the best, achieving near perfect AUROC. 

Given a vector of concept predictions $c$ and classification head vector corresponding to a class $w$, we compute the contribution of concept $i$ to the class as $\frac{w_i \times c_i}{\sum_j |w_j \times c_j|}$. We call these logit shares, as each score is the share of activation on a given logit from one concept divided by the sum of contributions to that logit over all concepts. 

Figure \ref{fig:cbn_eg_frog} and \ref{fig:cbn_extra_egs} shows how inference on a CBM is more interpretable, thanks to direct computation of logit shares with respect to a set of concepts. Specifically, we list the predicted class and the three concepts with largest logit shares for the predicted class. We additionally list the runner-up class, along with the concepts with the largest difference in activation shares for the predicted logit and the runner up logit. These concepts amount to ones that are much more influential for the predicted class than the runner up, and can provide insight as to why the predicted class was chosen over the runner up. This interpretability benefit can be particularly useful in understanding failure modes. For example, the rightmost panel in Figure \ref{fig:cbn_extra_egs} displays a \emph{truck} image misclassified as a \emph{ship}. The concept with the largest drop in logit share between the \emph{ship} and \emph{truck} logits is the color \emph{blue}. Arguably, this reveals a potentially problematic spurious correlation the CBM attributes to the \emph{ship} class, perhaps because of the blue water and blue sky that is often present in images of ships. Because the color blue is prominent in the truck image (from the sky), the model misclassifies it as \emph{ship}. Identifying spurious correlations is the first towards improving the robustness of models to changes in spurious features, and the CBM makes identifying such correlations (namely to human interpretable concepts) easier.

\section{Concept Logic for Image Retrieval}
\label{app-sec:concept_logic}
\begin{figure}[h]
    \centering
    \includegraphics[width=0.75\linewidth]{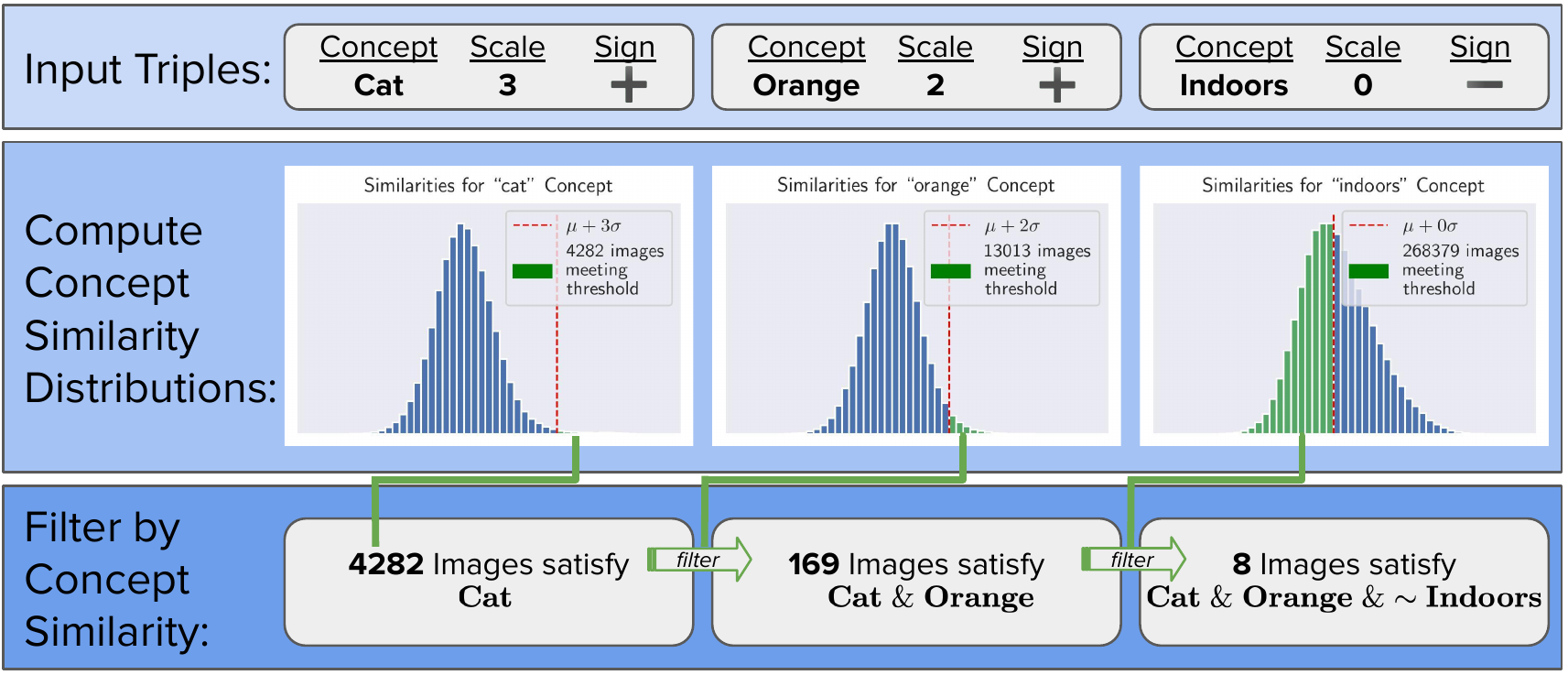}
    \caption{Overview of our concept logic image retrieval technique. Instead of directly searching for images of \emph{orange cats that are not indoors}, we decompose the query into concept constraints of varying degree. Our method grants the querier more control in their search, and also circumvents limitations of CLIP's text encoder when processing negations and longer queries.}
    \label{fig:concept_logic_diagram}
\end{figure}

\begin{table}[htb]
    \centering
    \begin{tabular}{c|c|c} \toprule
        Encoder & Concept Constraints & All Retrieved Images \\ \midrule
        \multirow{13}{*}{Dino-ResNet50} &  & \multirow{13}{*}{\includegraphics[width=0.4\textwidth]{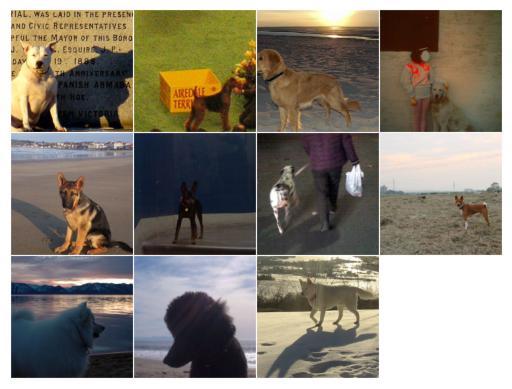}} \\
        & & \\
        & & \\
        & & \\
        & & \\
        & (`a dog', 2.25, 1) & \\
        & (`the beach', 2, 1) & \\
        & (`the sunset', 2, 1) & \\ 
        & & \\
        & & \\
        & & \\
        & & \\
        & & \\
        \midrule
        \multirow{9}{*}{Dino-ResNet50} &  & \multirow{9}{*}{\includegraphics[width=0.4\textwidth]{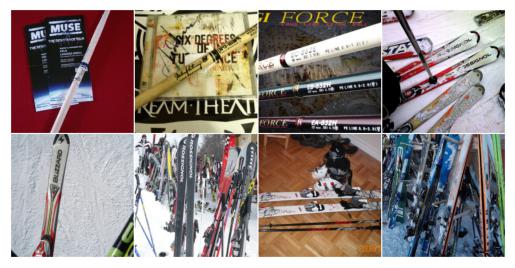}} \\
        & & \\
        & & \\
        & (`skis', 4, 1) & \\
        & (`snow', 2, -1) & \\
        & (`human', 1,-1) & \\ 
        & & \\
        & & \\
        & & \\
        \midrule
        \multirow{9}{*}{Standard ResNet50} &  & \multirow{9}{*}{\includegraphics[width=0.4\textwidth]{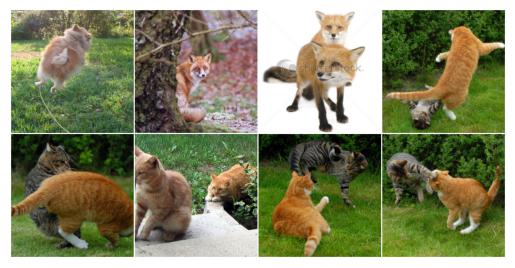}} \\
        & & \\
        & & \\
        & (`cat', 3, 1) & \\
        & (`orange', 2, 1) & \\
        & (`indoors', 0, -1) & \\ 
        & & \\
        & & \\
        & & \\
        \bottomrule
    \end{tabular}
    \caption{Complete details for Figure \ref{fig:concept_logic}, including concept constraints in the form used to query images and all retrieved images for a set of concept constraints and a fixed vision encoder.}
    \label{tab:full_concept_logic}
\end{table}

Concept Logic simply applies a set of filters to obtain images that meet several concept requirements. The constraints are of the form (concept, scale $k$, sign $s$). This means that the similarity of the aligned representation of an image to the text-to-concept vector must either be $k$ standard deviations above or below (depending on $s$) the mean similarity for that vector. The scale parameter allows for easy control of how strongly each concept in the query be should take into account. 

Figure \ref{fig:concept_logic_diagram} diagrams our approach, specifically for the example of retrieving images of \emph{orange cats that are not indoors}. For a set of concept constraints, we first encode them as input triples, indicating the severity with which each concept constraint should be applied. Reducing the severity of a constraint leads to the retrieval of more images, at the cost of including more erroneous instances. The sign of the constraint allows for negation of certain concepts (e.g. returning images of cats that are \emph{not} indoors). We observe CLIP's text encoder to struggle with negations; concept logic circumvents this issue by returning instances \emph{least} similar to a concept. 

For each concept constraint, we encode the concept and compute the similarity of all aligned representations from the data pool of interest to the concept vector, and then obtain a subset of images that satisfy the given constraint. Finally, we take the intersection of all returned subsets, resulting in a final set of retrieved images that satisfy all concept constraints.  

Table \ref{tab:full_concept_logic} outlines the complete details for the concept logic based image retrieval previewed in Figure \ref{fig:concept_logic}. In the figure, we present one of the retrieved images given a set of concept constraints and a fixed vision encoder. The table contains the exact concept constraints, as well as all images that the query returns. 

\section{Concept-to-text}
\label{app-sec:cct}

We now provide additional details on the concept-to-text experiment of Section \ref{sec:cct}. The method to perform concept-to-text immediately follows from our work aligning diverse model spaces to CLIP, and the work of \citet{zerocap} (ZeroCap), which decodes CLIP vectors to text with no training required. Our use case is novel compared to those explored in ZeroCap, as we decode \emph{general} vectors in a feature space, as opposed to representations directly corresponding to natural images.

Namely, we decode \emph{classification head vectors} from three pretrained ImageNet classifiers (Swin transformer, standard supervised ResNet-50, and Dino-trained ViT) of roughly equivalent size. Recall that the predicted class of an image is the argmax of the product of classification head vectors and the image's representation (plus a bias vector). Thus, we would expect each classification vector to correspond to a concepts relevant to the class, such as the class object itself, similar object, or things that frequently co-occur with the class object.  

We make one small modification to classification head vectors before decoding them. While they exist in the same space as image representations, their norm conceivably may be much different than the norms of image representations, on which our aligners are trained. So long as all classification head vectors have similar norm, the argmax of the product of classification head vectors and an image representation would still be effective in predicting said image's class. However, this would result in poor alignment, as the classification head vectors may be out of distribution for the linear aligner. The solution to this problem is simple: we rescale all classification head vectors by a constant so that the variance over all of their elements is equal to the variance over all representations of images used to train the aligner. This minor modification drastically improved performance, particularly for the transformer-based models we studied. 

\begin{figure*}
    \centering
    \includegraphics[width=0.9\linewidth]{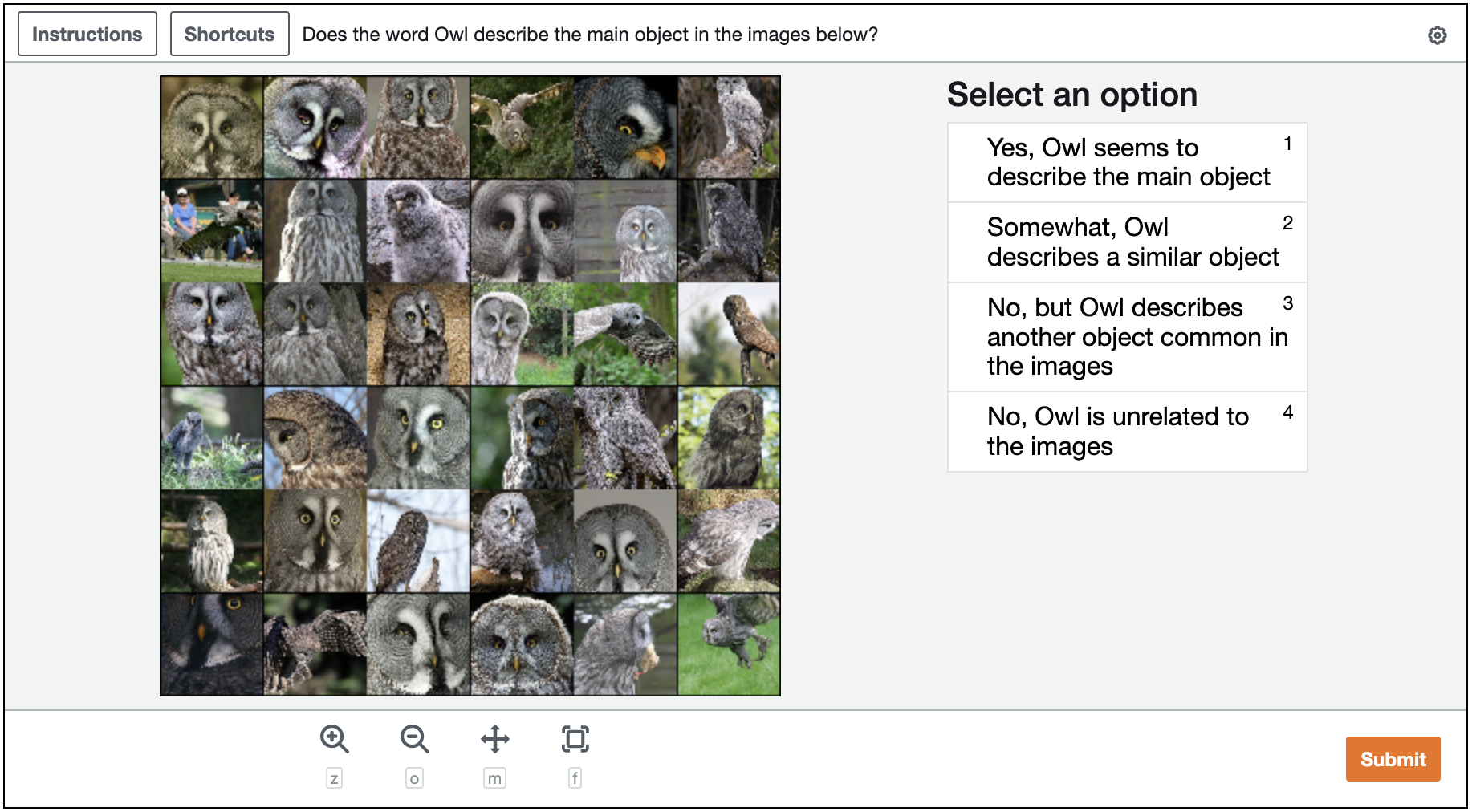}
    \caption{Sample HIT shown to MTurk workers. The decoded caption is `owl', and the corresponding class name is `grey owl'.}
    \label{app-fig:mturk}
\end{figure*}

To recap, our procedure to decode a classification head vector is to (i) rescale it, (ii) align to CLIP using a linear aligner, and (iii) decode using ZeroCap, with the prompt of ``Image of a '' and a target output sequence length of $1$. We choose a sequence length of $1$ so to reduce hallucinations from GPT-2. All other hyperparameters for ZeroCap are set to the default captioning settings). We note this process is extremely efficient: in minutes, we can decode all one thousand classification head vectors for a given model. 

\begin{table}[htb]
    \centering
    \begin{tabular}{c|ccc} \toprule
          & Describes Concept Relevant &
         Describes Concept Similar &
         Describes Main Object \\ 
         Model & to Images from Class & to Main Object & \\ \midrule
        Swin (S) & 94.48\% & 84.60\% & 69.51\% \\ 
        ResNet-50 & 95.14\% & 88.37\% & 71.36\%  \\ 
        Dino ViTs8 & 92.18\% & 76.45\% & 60.47\%  \\ \hline
        {\bf Average} & {\bf 93.93\%} & {\bf 83.14\%} & {\bf 67.11\%}  \\ \bottomrule
    \end{tabular}
    \caption{Complete results from human study assessing quality of using concept-to-text to label classification head vectors from various ImageNet trained models. We list the rate for which the decoded text satisfies some relation (e.g. `describes main object') to the collage of sample images shown for a given class.}
    \label{tab:full_cct}
\end{table}

\begin{table}[h]
    \centering
    \begin{tabular}{c|ccc} \toprule
         & Related vs.  & Similar vs. Dissimilar & Main Object vs. Similar \\
         Model & Unrelated & vs. Unrelated & vs. Dissimilar vs. Unrelated \\ \midrule
        Swin (S) & 89.77\% & 76.63\% & 56.07\% \\ 
        ResNet-50 & 91.27\% & 82.25\% & 56.87\% \\ 
        Dino ViTs8 & 85.77\% & 67.13\% & 48.50\% \\ \hline
        Average & 88.94\% & 75.34\% & 53.81\% \\ 
        Random & 25\% & 11.1\% & 6.25\% \\ \bottomrule
    \end{tabular}
    \caption{Inter-annotator agreement, at various grains. \emph{Random} lists the agreement expected between two random annotators, computed as $1/n$ where $n$ refers to the number of possible choices for a given metric. We use two annotators for every query.}
    \label{tab:cct_iaa}
\end{table}

\looseness=-1
To assess the quality of our decodings, we use a human study conducted on Amazon Mechanical Turk. Figure \ref{app-fig:mturk} displays a screenshot for a single task, where a human compares a collage of images from a given class to the single word decoded for the corresponding classification head vector. Specifically, for the question `does the word \{\} describe the main object in the images below?', annotators choose between four responses: (i) yes, \{\} seems to describe the main object, (ii) somewhat, \{\} describes a similar object, (iii) no, but \{\} describes another object common in the images, (iv) no, \{\} is unrelated to the images.   

The results of our human study are presented in Table \ref{tab:full_cct}, structured as follows: The first column is the rate that humans picked any option aside from the fourth option. The second column is the rate that humans picked the first or second option. The third column is the rate that the humans picked the first option. Also, we provide sample decodings for $45$ randomly selected ImageNet classes in table \ref{tab:cct_samples}. As shown in the main text, decoded concepts are almost always relevant to images from the class. Interestingly, in about $10\%$ of cases on average (obtained via subtracting column 1 from column 2), the decoded word describes a common object in the images that is distinct from the main object. These cases correspond to when the decoded word is a spurious feature to the class (e.g. `freeway' for the class 'Water Tower'). Identifying spurious features is a potential application of concept-to-text. 

{\bf Additional logistical details of study:} For each model-class pair, we obtain two responses. MTurk workers are compensated $\$0.05$ (USD) per task, resulting to an average rate of $\$15$ per hour. We compute provide inter-annotator agreement for our human study in table \ref{tab:cct_iaa}. Inter annotator agreement is generally high. 



\begin{table}[htb]
    \centering
    \begin{tabular}{c|ccc} \toprule
        & \multicolumn{3}{c}{Model} \\
        Class & Swin (S) & ResNet-50 & Dino ViTs8 \\ \midrule
American Alligator &  Lizard &  lizard &  Florida \\ \hline
Messenger Bag &  patch &  pocket &  tet \\ \hline
Spindle &  spinning &  spinning &  coral \\ \hline
Radio &  radio &  radio &  telephone \\ \hline
Ipod &  screenshot &  Nokia &  USB \\ \hline
Yorkshire Terrier &  photo &  puppy &  red \\ \hline
Hourglass &  clock &  clock &  clock \\ \hline
Lion &  lion &  Lion &  lion \\ \hline
Revolver &  handgun &  handgun &  Glock \\ \hline
Scoreboard &  sign &  billboard &  game \\ \hline
Wallaby &  deer &  bunny &  shrew \\ \hline
Tent &  Tibetan &  cave &  tent \\ \hline
Monastery &  monastery &  monastery &  monastery \\ \hline
Front Curtain &  small &  Pluto &  world \\ \hline
Golf Ball &  a &  golf &  golf \\ \hline
Notebook Computer &  laptop &  Chromebook &  notebook \\ \hline
Water Tower &  large &  tower &  freeway \\ \hline
Gas Pump &  vending &  garage &  garage \\ \hline
Smooth Newt &  Lizard &  lizard &  slime \\ \hline
Platypus &  mole &  mole &  crocod \\ \hline
Paintbrush &  painting &  paint &  pen \\ \hline
Product Packet / Packaging &  packet &  is &  Saturn \\ \hline
Chiffonier &  replica &  box &  closet \\ \hline
Water Jug &  jug &  jug &  cup \\ \hline
Boa Constrictor &  python &  python &  python \\ \hline
Rapeseed &  field &  yellow &  farm \\ \hline
Police Van &  police &  van &  police \\ \hline
Maltese &  replica &  puppy &  Pluto \\ \hline
Pot Pie &  previously &  clam &  pan \\ \hline
Menu &  menu &  menu &  strawberry \\ \hline
Red Wine &  red &  red &  red \\ \hline
Mosquito Net &  young &  bedroom &  bedroom \\ \hline
Poncho &  swarm &  condom &  square \\ \hline
Basenji &  building &  dog &  puppies \\ \hline
Turnstile &  human &  gate &  hospital \\ \hline
Sea Slug &  coral &  crocod &  squid \\ \hline
Computer Keyboard &  keyboard &  keyboard &  computer \\ \hline
Ballpoint Pen &  3 &  glucose &  neuron \\ \hline
Plate Rack &  wall &  table &  table \\ \hline
Bridegroom &  wedding &  wedding &  flower \\ \hline
Fire Salamander &  pair &  Lizard &  frog \\ \hline
T-Shirt &  member &  shirt &  comet \\ \hline
Eastern Diamondback Rattlesnake &  python &  snake &  python \\ \hline
Fiddler Crab &  crabs &  crabs &  crab \\ 
        \bottomrule
        \end{tabular}
    \caption{Decodings (obtained via concept-to-text) of classification head vectors for $45$ randomly selected ImageNet classes from three pretrained models. The majority of decoded words are similar (though often broader) to the corresponding class object.}
    \label{tab:cct_samples}
\end{table}

\section{Limitations}
\label{app-sec:limitations}

We note that the concept vectors we find are not always perfect.
However, more refined concept vectors can be found by
(1) better prompt engineering and
(2) extracting more relevant image samples to that concept.

The second point can be achieved by inspecting the most and least similar images retrieved for a desired concept, and removing erroneous examples. Then, one can obtain a concept vector in the ordinary manner (i.e. training a binary classifier in representation space to separate positive and negative examples of the concept), or more simply by taking the average of their encoded representations. 

Finally, we note that our method generally becomes better when any of the components which are involved in concept-to-text and text-to-concept procedures become better.
Indeed, better CLIP models, more powerful vision encoders, and improved generative language models can contribute to an improved performance of concept-to-text and text-to-concept.

\section{Additional Related Works}

Some recent works also investigate bridging image and text models \cite{merullo2022linearly, lit, frozen}, though their training mechanisms typically involve text supervision and propagating through both image and text backbones, which can make them much more intensive. Previous efforts have sought to map image spaces to semantic using cycle consistency objectives for zero-shot learning \cite{gzsl} or image translation \cite{unit}. 

Our method of aligner training using representations from a pretrained model draws some parallels to knowledge distillation \cite{Hinton2015DistillingTK}, where activations from a more powerful model are utilized in training a smaller one to behave similarly \cite{Beyer2021KnowledgeDA}. However, the crucial difference between our work and standard knowledge distillation in that in our method, the vision encoders we align to CLIP remain fixed. We do not wish to distill the knowledge of CLIP to other models – in fact, we intentionally fix the off-the-shelf model and only allow for a minimal transformation (affine) of its representation space. Instead, we argue that existing vision models already encode many human concepts in their feature space, though accessing this information is challenging without a text encoder that maps to the same space. Our method allows for interpretation of the off-the-shelf model’s space in an efficient and flexible way; i.e. by obtaining concept activation vectors (CAVs) directly from text using CLIP’s text encoder. In summary, since the aim of our work is interpretability, we do not wish to transfer or distill knowledge from one model to another. Rather, we seek to allow existing models to work with one another in an efficient manner. We hope our work inspires others to investigate ways specialized models can be interfaced together to accomplish novel ends in inexpensive ways.   

\end{document}